\documentclass[10pt,twocolumn,letterpaper]{article}

\usepackage{iccv}
\usepackage{times}
\usepackage{epsfig}
\usepackage{graphicx}
\usepackage{amsmath}
\usepackage{amssymb}

\usepackage{graphicx}
\usepackage{comment}
\usepackage{lipsum}   %
\usepackage{caption}
\usepackage{subcaption}
\usepackage{float}

\usepackage{color,xcolor}
\usepackage{epsfig}

 \usepackage{microtype} 

\usepackage{chngcntr}

\usepackage{tabularx}
\usepackage{adjustbox}
\usepackage{array}
\usepackage{booktabs}
\usepackage{colortbl}
\usepackage{float,wrapfig}
\usepackage{hhline}
\usepackage{multirow}
\usepackage{subcaption}

\usepackage{amsmath,amsfonts,amssymb} %
\usepackage{bm}
\usepackage{nicefrac}
\usepackage{microtype}
\usepackage{dsfont}
\usepackage{changepage}
\usepackage{extramarks}
\usepackage{fancyhdr}
\usepackage{lastpage}
\usepackage{setspace}
\usepackage{soul}
\usepackage{xspace}

\usepackage{paralist}
\usepackage{enumitem}

\usepackage[font=small]{caption}

\usepackage{fp}
\usepackage{expl3}[2012-07-08]
\ExplSyntaxOn
\cs_new_eq:NN \fpeval \fp_eval:n
\ExplSyntaxOff

\usepackage{amsmath,amsfonts,bm}

\def\x{{x}}

\def\xi{{\x_i}}

\newcommand{\reffig}[1]{Figure~\ref{fig:#1}}
\newcommand{\refsec}[1]{Section~\ref{sec:#1}}

\newcommand{\reftbl}[1]{Table~\ref{tbl:#1}}

\newcommand{\lblfig}[1]{\label{fig:#1}}
\newcommand{\lblsec}[1]{\label{sec:#1}}
\newcommand{\lbleq}[1]{\label{eq:#1}}
\newcommand{\lbltbl}[1]{\label{tbl:#1}}

\newcommand{\ignorethis}[1]{}

\newcommand{\myparagraph}[1]{\smallskip \noindent \textbf{#1}}

\def\eqref#1{equation~\ref{#1}}

\def\1{\bm{1}}

\def\eps{{\epsilon}}

\def\vc{{\bm{c}}}
\def\vd{{\bm{d}}}

\def\vr{{\bm{r}}}

\def\vx{{\bm{x}}}

\def\vz{{\bm{z}}}

\DeclareMathAlphabet{\mathsfit}{\encodingdefault}{\sfdefault}{m}{sl}
\SetMathAlphabet{\mathsfit}{bold}{\encodingdefault}{\sfdefault}{bx}{n}

\newcolumntype{L}[1]{>{\raggedright\let\newline\\\arraybackslash\hspace{0pt}}m{#1}}
\newcolumntype{C}[1]{>{\centering\let\newline\\\arraybackslash\hspace{0pt}}m{#1}}
\newcolumntype{R}[1]{>{\raggedleft\let\newline\\\arraybackslash\hspace{0pt}}m{#1}}

\newcommand{\ignore}[1]{}

\makeatletter
\DeclareRobustCommand\onedot{\futurelet\@let@token\@onedot}
\def\@onedot{\ifx\@let@token.\else.\null\fi\xspace}
\def\eg{e.g\onedot,\xspace}

\def\etal{\emph{et al}\onedot}
\makeatother

\definecolor{MyDarkBlue}{rgb}{0,0.08,1}
\definecolor{MyDarkGreen}{rgb}{0.02,0.6,0.02}
\definecolor{MyDarkRed}{rgb}{0.8,0.02,0.02}
\definecolor{MyDarkOrange}{rgb}{0.40,0.2,0.02}
\definecolor{MyPurple}{RGB}{111,0,255}
\definecolor{MyRed}{rgb}{1.0,0.0,0.0}
\definecolor{MyGold}{rgb}{0.75,0.6,0.12}
\definecolor{MyDarkgray}{rgb}{0.66, 0.66, 0.66}
\definecolor{myorange}{RGB}{255,69,0}

\newcommand{\newtext}{\textcolor{black}}
\newcommand{\newnewtext}{\textcolor{black}}

\usepackage[pagebackref=true,breaklinks=true,letterpaper=true,colorlinks,bookmarks=false]{hyperref}

\iccvfinalcopy %

\def\iccvPaperID{0000} %

\begin{document}

\title{Editing Conditional Radiance Fields}    

\author{Steven Liu\textsuperscript{1} \enskip Xiuming Zhang\textsuperscript{1} \enskip Zhoutong Zhang\textsuperscript{1} \enskip Richard Zhang\textsuperscript{2} \enskip Jun-Yan Zhu\textsuperscript{2,3} \enskip Bryan Russell\textsuperscript{2}
\\
\textsuperscript{1}MIT \qquad \textsuperscript{2}Adobe Research \qquad \textsuperscript{3}CMU}

\maketitle

\begin{abstract}
   \newtext{A neural radiance field (NeRF) is a scene model supporting high-quality view synthesis, optimized per scene. In this paper, we explore enabling user editing of a category-level NeRF -- also known as a conditional radiance field -- trained on a shape category. Specifically, we introduce a method for propagating coarse 2D user scribbles to the 3D space, to modify the color or shape of a local region.}
First, we propose a conditional radiance field that incorporates new modular network components, including a shape branch that is shared across object instances.
Observing multiple instances of the same category, our model learns underlying part semantics without any supervision, thereby allowing the propagation of coarse 2D user scribbles to the entire 3D region (\eg chair seat). 
Next, %
we propose a hybrid network update strategy that targets specific network components, which balances efficiency and accuracy.  
During user interaction, we formulate an optimization problem that both satisfies the user's constraints and preserves the original object structure.
We demonstrate our approach on various editing tasks over three shape datasets and show that it outperforms prior neural editing approaches. 
Finally, we edit the appearance and shape of a real photograph and show that the edit propagates to extrapolated novel views.

\end{abstract}

\section{Introduction}
\lblsec{intro}

\begin{figure}
  \centering
  \includegraphics[width=1.\linewidth]{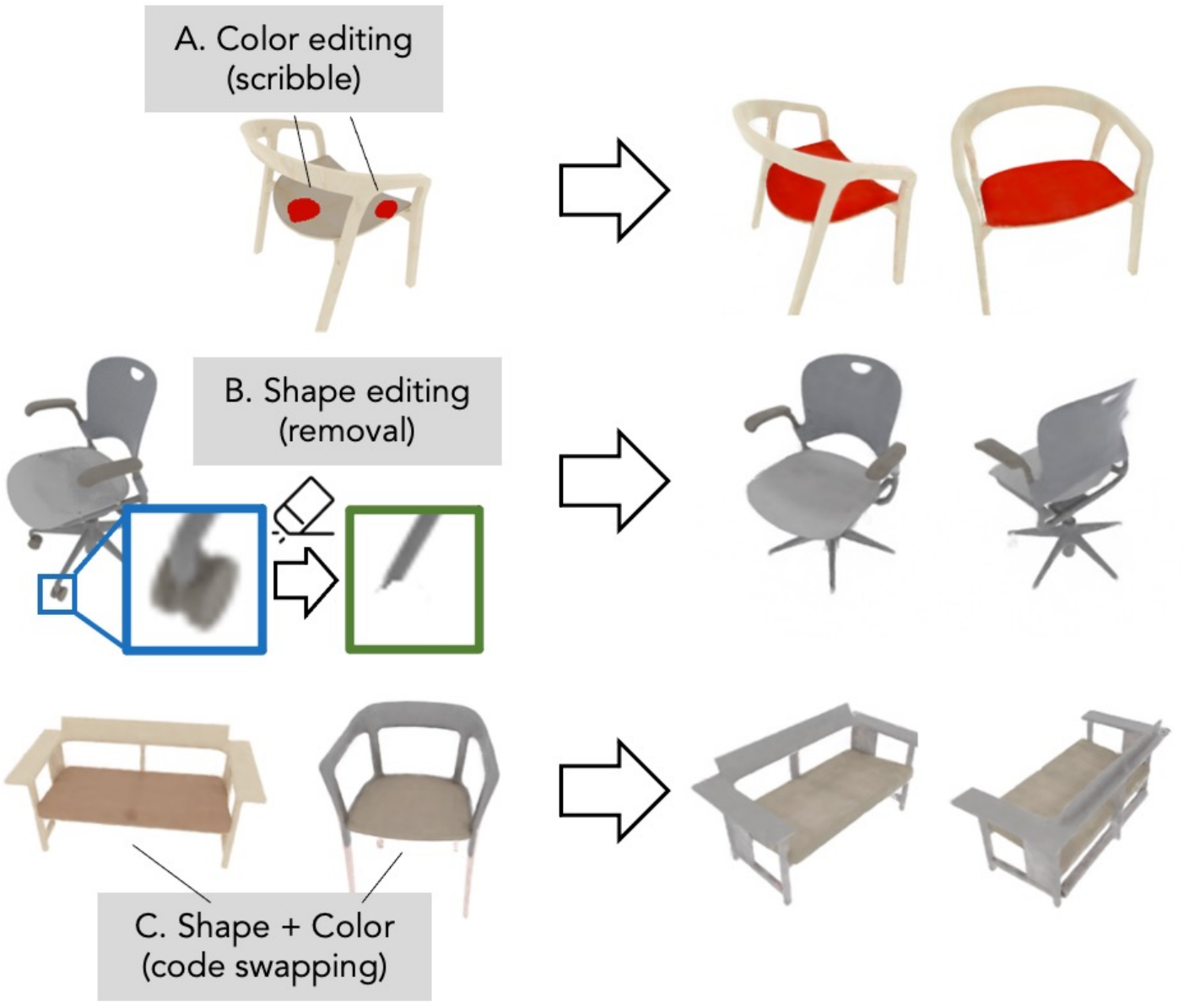}
  \caption{\newtext{{\bf Editing a conditional radiance field.}  Given a conditional radiance field trained over a class of objects, we demonstrate three editing applications:}
  (A) color editing, (B) shape editing, and (C) color/shape transfer.
  A user provides coarse scribbles over a local \newtext{region} of interest or selects a target object instance. \newtext{Local} edits propagate to the desired region in 3D and are consistent across different rendered views.
  }
  \vspace{-20pt}
  \lblfig{teaser}
\end{figure}

3D content creation often involves manipulating high-quality 3D assets for visual effects or augmented reality applications, and part of a 3D artist's workflow consists of making local adjustments to a 3D scene's appearance and shape~\cite{3dstutorial,blendertutorial}.  
Explicit representations give artists control of the different elements of a 3D scene. 
For example, the artist may use mesh processing tools to make local adjustments to the scene geometry or change the surface appearance by manipulating a texture atlas~\cite{schmidt14star}. 
In an artist's workflow, such explicit representations are often created by hand or procedurally generated.

While explicit representations are powerful, there remain significant technical challenges in automatically acquiring a high-quality explicit representation of a real-world scene due to view-dependent appearance, complex scene topology, and varying surface opacity. 
Recently, implicit continuous volumetric representations have shown high-fidelity capture and rendering of a variety of 3D scenes and overcome many of the aforementioned technical challenges~\cite{park2019deepsdf,sitzmann2019srns,mildenhall2020nerf,Riegler2020FVS,schwarz2020graf}. 
Such representations encode the captured scene in the weights of a neural network. The neural network learns to render view-dependent colors from point samples along cast rays, with the final rendering obtained via alpha compositing~\cite{porter1984alpha}. 
This representation enables many photorealistic view synthesis applications~\cite{martinbrualla2020nerfw,niemeyer2020dvr}. %
However, we lack critical knowledge in how to enable artists' control and editing in this representation. 

Editing an implicit continuous volumetric representation is challenging. 
First, how can we effectively propagate sparse 2D user edits to fill the entire corresponding 3D region in this representation? 
Second, the neural network for an implicit representation has millions of parameters. It is unclear which parameters control the different aspects of the rendered shape and how to change the parameters according to the sparse local user input. 
While prior work for 3D editing primarily focuses on editing an explicit representation~\cite{schmidt14star}, they do not apply %
to neural representations.

In this paper, we study how to enable users to edit and control an implicit continuous volumetric representation of a 3D object. 
As shown in \reffig{teaser}, we consider three types of user edits: (i) changing the appearance of a local part to a new target color (\eg changing the chair seat's color from beige to red), (ii) modifying the local shape (\eg removing a chair's wheel or swapping in new arms from a different chair), and (iii) transferring the color or shape from a target object instance. 
The user performs 2D local edits by scribbling over the desired location of where the edit should take place and selecting a target color or local shape.

We address the challenges in editing an implicit continuous representation by investigating how to effectively update a conditional radiance field to align with a target local user edit. We make the following contributions. 
First, we learn a conditional radiance field over an entire object class to model a rich prior of plausible-looking objects. Unexpectedly, this prior often allows the propagation of sparse user scribble edits to fill a selected region. We demonstrate complex edits without the need to impose explicit spatial or boundary constraints. 
Moreover, the edits appear consistently when the object is rendered from different viewpoints. 
Second, to more accurately reconstruct shape instances, we introduce a shape branch in the conditional radiance field that is shared across object instances, which implicitly biases the network to encode a shared representation whenever possible. 
Third, we investigate which parts of the conditional radiance field's network affect different editing tasks. We show that shape and color edits can effectively take place in the later layers of the network. This finding motivates us to only update these layers and enables us to produce effective user edits with significant computational speed-up. 
Finally, we introduce color and shape editing losses to satisfy the user-specified targets, while preserving the original object structure.

We demonstrate results on three shape datasets with varying levels of appearance, shape, and training view complexity. We show the effectiveness of our approach for object view synthesis as well as color and shape editing, compared to prior neural editing methods.  
Moreover, we show that we can edit the appearance and shape of a real photograph and that the edit propagates to extrapolated novel views.  
 We highly encourage viewing our \href{https://youtu.be/9qwRD4ejOpw}{video} to see our editing demo in action. Code and more results are available at our \href{https://github.com/stevliu/editnerf}{GitHub} repo and \href{http://editnerf.csail.mit.edu/}{website}.

\section{Related Work}
\lblsec{related}
Our work is related to novel view synthesis and interactive appearance and shape editing, which we review here.

\myparagraph{Novel view synthesis.} 
Photorealistic view synthesis has a storied history in computer graphics and computer vision, which we briefly summarize here. 
The goal is to infer the scene structure and view-dependent appearance given a set of input views. 
Prior work reasons over an explicit~\cite{buehler2001lumigraph,debevec1996facade,groueix2018atlasnet,wood2000surface} or discrete volumetric~\cite{flynn2019deepview,kutulakos2000carving,li2020crowdsampling,lombardi2019neuralvolumes,mildenhall2019llff,penner2019soft3d,seitz1999coloring,sitzmann2019deepvoxels,srinivasan19mpi,szeliski1999stereo,zhou2018stereo,zhu2018von} representation of the underlying geometry. 
However, both have fundamental limitations -- explicit representations often require fixing the structure's topology and have poor local optima, while discrete volumetric approaches scale poorly to higher resolutions. 

Instead, several recent approaches implicitly encode a continuous volumetric representation of shape~\cite{chen2018implicit_decoder,Genova_2020_CVPR,liu2019implicit,liu2020dist,mescheder2019occupancy,Michalkiewicz_2019_ICCV,OechsleICCV2019,park2019deepsdf,peng2020convocc,saito2020pifuhd} or both shape and view-dependent appearance~\cite{martinbrualla2020nerfw,mildenhall2020nerf,niemeyer2020dvr,Riegler2020FVS,schwarz2020graf,sitzmann2019srns,tancik2020meta,yu2020pixelnerf,chanmonteiro2020pi-GAN,xian2020space} in the weights of a neural network. 
These latter approaches overcome the aforementioned limitations and have resulted in impressive novel-view renderings of complex real-world scenes. 
\newtext{Closest to our approach is Schwarz \etal~\cite{schwarz2020graf,chanmonteiro2020pi-GAN}, where they build a generative radiance field over an object class and include latent vectors for the shape and appearance of an instance. Different from their methods, we
include an instance-agnostic branch in our neural network, which inductively biases the network to capture common features across the shape class. As we will demonstrate, this inductive bias more accurately captures the shape and appearance of the class. %
Moreover, we do not require an adversarial loss to train our network and instead optimize a photometric loss, which allows our approach to directly align to a single view of a novel instance. 
Finally, our work is the first to address the question of how to enable a user to make local edits in this new representation.}

\myparagraph{Interactive appearance and shape editing.}
There has been much work on interactive tools for selecting and cloning regions~\cite{agarwala2004interactive,li2004lazy,perez2003poisson,rother2004grabcut} and editing single still images~\cite{an2008appprop,avidan2007seam,barnes2009patchmatch,levin2004colorization}. Recent works have focused on integrating user interactions into deep networks either through optimization~\cite{abdal2019image2stylegan,bau2019ganpaint,zhu2016generative,brock2017neural} or a feed-forward network with user-guided inputs~\cite{zhang2017real,Faceshop,park2020swapping,olszewski2020intuitive}.
Here, we are concerned with editing 3D scenes, which has received much attention in the computer graphics community. 
Example interfaces include 3D shape drawing and shape editing using inflation heuristics~\cite{igarashi1999teddy}, stroke alignment to a depicted shape~\cite{chen2013threesweep}, and learned volumetric prediction from multi-view user strokes~\cite{delanoy2018sketching}. 
There has also been work to edit the appearance of a 3D scene, \eg via transferring multi-channel edits to other views~\cite{hennessey2017transfer}, scribble-based material transfer~\cite{an2011appwarp}, editing 3D shapes in a voxel representation~\cite{liu2017interactive}, and relighting a scene with a paint brush interface~\cite{Pellacini:2007:LWP}. 
\newtext{Finally, there has been work on editing light fields~\cite{horn2007lightshop,jarabo2014lightfields,jarabo2011efficientpropagation}.} 
We encourage the interested reader to review this survey on artistic editing of appearance, lighting, and material~\cite{schmidt14star}. 
These prior works operate over \newtext{light fields or explicit/discrete volumetric geometry} whereas we seek to incorporate user edits in learned implicit continuous volumetric representations. 

A closely related concept is edit propagation~\cite{an2008appprop,endo2016deepprop,hasinoff2010search,xu2009efficient,yucer2012transfusive}, which propagates sparse user edits on a single image to an entire photo collection or video. In our work, we aim to propagate user edits to  volumetric data for rendering under different viewpoints. 
Also relevant is recent work on applying local ``rule-based'' edits to a trained generative model for images~\cite{bau2020rewriting}. 
We are inspired by the above approaches and adapt it to our new 3D neural editing setting.  %

\section{Editing a Conditional Radiance Field}

\newcommand{\colorout}{\hat{C}}
\newcommand{\ray}{\vr}
\newcommand{\shapecode}{\vz^{(s)}}
\newcommand{\colorcode}{\vz^{(c)}}
\newcommand{\network}{\mathcal{F}}

\begin{figure}
  \centering
  \includegraphics[width=\linewidth]{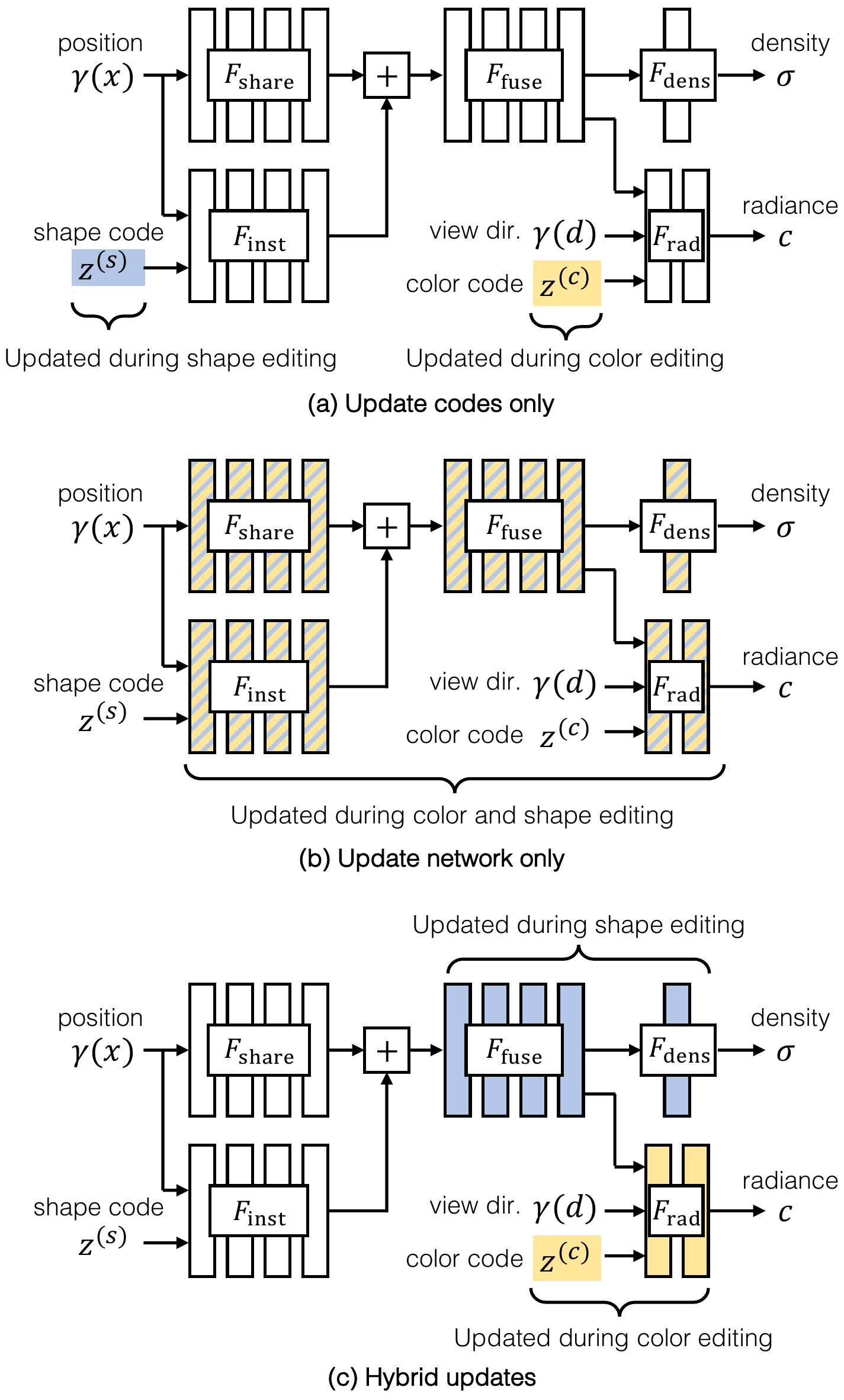}
  \caption{
  {\bf Conditional radiance field network.} Our network maps a 3D location $\vx$, viewing direction $\vd$, and instance-specific shape code $\shapecode$ and color code $\colorcode$ to radiance $\vc$ and scalar density $\sigma$.  
The network is composed of modular parts for better shape and color disentanglement. 
  We train our network over a collection of 3D objects (\refsec{learning}). \newtext{As highlighted, only a subset of the network components need to be updated during editing (\refsec{modular})}.
  }
  \vspace{-10pt}
  \lblfig{arch}
\end{figure}

Our goal is to allow user edits of a continuous volumetric representation of a 3D scene. 
In this section, we first describe a new neural network architecture that more accurately captures the shape and appearance of an object class. We then describe how we update network weights to achieve color and shape editing effects. 

To achieve this goal, we build upon the recent neural radiance field (NeRF) representation~\cite{mildenhall2020nerf}. 
While the NeRF representation can render novel views of a particular scene, we seek to enable editing over an entire shape class, \eg ``chairs''. 
For this, we learn a conditional radiance field model that extends the NeRF representation with latent vectors over shape and appearance.  
The representation is trained over a set of shapes belonging to a class, and each shape instance is represented by latent shape and appearance vectors. 
The disentanglement of shape and appearance allows us to modify certain parts of the network during editing. 

Let $\vx = (x, y, z)$ be a 3D location, $\vd = (\phi, \theta)$ be a viewing direction, and $\shapecode$ and $\colorcode$ be the latent shape and color vectors, respectively. 
Let $\left(\vc, \sigma\right) = \network{\left(\vx, \vd, \shapecode, \colorcode\right)}$ be the neural network for a conditional radiance field that returns a radiance $\vc = (r, g, b)$ and a scalar density $\sigma$. 
The network $\network$ is parametrized as a multi-layer perceptron (MLP) such that the density output $\sigma$ is independent of the viewing direction, while the radiance $\vc$ depends on both position and viewing direction. 

To obtain the color at a pixel location for a desired camera location, first, $N_c$ 3D points $\{ t_i \}_{i=1}^{N_c}$ are sampled along a cast ray $\ray$ originating from the pixel location (ordered from near to far). 
Next, the radiance and density values are computed at each sampled point with network $\network$. 
Finally, the color is computed by the ``over'' compositing operation~\cite{porter1984alpha}. 
Let $\alpha_i = 1-\exp{\left(-\sigma_i \delta_i\right)}$ be the alpha compositing value of sampled point $t_i$ and $\delta_i=t_{i+1}-t_i$ be the distance between the adjacent sampled points. 
The compositing operation, which outputs pixel color $\colorout$, is the weighted sum:
\begin{equation}
\colorout{\left(\ray,\shapecode,\colorcode\right)} = \sum_{i=1}^{N_c-1} c_i \alpha_i \exp{\left(-\sum_{j=1}^{i-1} \sigma_j \delta_j\right)}.
\label{eqn:composite}
\end{equation}
Next, we describe details of our network architecture and our training and editing procedures.

\subsection{Network with Shared Branch}
\lblsec{learning}

\newtext{NeRF~\cite{mildenhall2020nerf} finds the inductive biases provided by positional encodings and stage-wise network design critical. Similarly, we find the architectural design choices important and aim for a modular model, providing an inductive bias for shape and color disentanglement.}
\newtext{These design choices allow for selected submodules to be finetuned during user editing (discussed further in the next section), enabling more efficient downstream editing. We illustrate our network architecture $\network$ in \reffig{arch}.}

\newtext{First, we learn a category-specific geometric representation with a \textit{shared shape network} $\network_\text{share}$ that only operates on the input positional encoding $\gamma(\vx)$~\cite{mildenhall2020nerf,vaswani2017attention}.  To modify the representation for a specific shape, an \textit{instance-specific shape network} $\network_\text{inst}$ is conditioned on both the shape code $\shapecode$ and input positional encoding. The representations are added and modified by a \textit{fusion shape network} $\network_\text{fuse}$.}
To obtain the density prediction $\sigma$, the output of $\network_\text{fuse}$ is passed to a linear layer, the \textit{output density network} $\network_\text{dens}$. To obtain the radiance prediction $\vc$, the output of $\network_\text{fuse}$ is concatenated with the color code $\colorcode$ and encoded viewing direction $\gamma(\vd)$ and passed through a two-layer MLP, the \textit{output radiance network} $\network_\text{rad}$. \newtext{We follow Mildenhall \etal~\cite{mildenhall2020nerf} for training and jointly optimize the latent codes via backpropagation through the network. We provide additional training details in the appendix.}

\subsection{Editing via Modular Network Updates}
\lblsec{modular}

We are interested in editing an instance encoded by our conditional radiance field. Given a rendering by the network $\network$ with shape $\shapecode_k$ and color $\colorcode_k$ codes, we desire to modify the instance given a set of user-edited rays.
We wish to optimize a loss $\mathcal{L_{\textrm{edit}}}(\network, \shapecode_k, \colorcode_k)$
over the network parameters and learned codes. 

Our first goal is to conduct the edit accurately -- the edited radiance field should render views of the instance that reflect the user's desired change. Our second goal is to conduct the edit efficiently. Editing a radiance field is time-consuming, as modifying weights requires dozens of forward and backward calls. Instead, the user should receive interactive feedback on their edits. To achieve these two goals, we consider the following strategies for selecting which parameters to update during editing. 

\myparagraph{Update the shape and color codes.}
One approach to this problem is to only update the latent codes of the instance, as illustrated in \reffig{arch}(a). While optimizing such few parameters leads to a relatively efficient edit, as we will show, this method results in a low-quality edit. 

\myparagraph{Update the entire network.} 
Another approach is to update all weights of the network, shown in \reffig{arch}(b). As we will show, this method is slow and can lead to unwanted changes in unedited regions of the instance. 

\myparagraph{Hybrid updates.} 
Our proposed solution, shown in \reffig{arch}(c), achieves both accuracy and efficiency by updating specific layers of the network. To reduce computation, we finetune the later layers of the network only. These choices speed up the optimization by only computing gradients over the later layers instead of over the entire network. When editing colors, we update only
$\network_\text{rad}$ and $\colorcode$ in the network, which reduces optimization time by $3.7\times$ over optimizing the whole network (from $972$ to $260$ seconds). When editing shape, we update only  $\network_\text{fuse}$ and  $\network_\text{dens}$, which reduces optimization time by $3.2 \times$ (from $1{,}081$ to $342$ seconds). 

In \refsec{shapeedits}, we further quantify the tradeoff between edit accuracy and efficiency.
To further reduce computation, we take two additional steps during editing. 

\myparagraph{Subsampling user constraints.} \newtext{During training, we sample a small subset of user-specified rays.
We find that this choice allows optimization to converge faster, as the problem size becomes smaller.} \newtext{For editing color, we randomly sample $64$ rays and for editing shape, we randomly sample a subset of $8{,}192$ rays.} \newtext{With this method, we obtain $24\times$ speedups for color edits and $2.9\times$ speedups for shape edits. Furthermore, we find that subsampling user constraints preserves edit quality; please refer to the appendix for additional discussion.} %

\myparagraph{Feature caching.}  NeRF rendering can be slow, especially when the rendered views are high-resolution. To optimize view rendering during color edits, we cache the outputs of the network that are unchanged during the edit. Because we only optimize $\network_\text{rad}$ during color edits, the input to $\network_\text{rad}$ is unchanged during editing. Therefore, we cache the input features for each of the views displayed to the user to avoid unnecessary computation. This optimization reduces the rendering time for a $256 \times 256$ image by $7.8 \times$ (from $6.2$ to under $0.8$ seconds). 

\newnewtext{
We also apply feature caching during optimization for shape and color edits. Similarly, we cache the outputs of the network that are unchanged during the optimization process to avoid unnecessary computation. Because the set of training rays is small during optimization, this caching is computationally feasible. %
We accelerate color edits by $3.2\times$ and shape edits by $1.9\times$.%
}

\subsection{Color Editing Loss}
In this section, we describe how to perform color edits with our conditional radiance field representation. 
To edit the color of a shape instance's part, the user selects a desired color and scribbles a foreground mask over a rendered view indicating where the color should be applied. 
The user may optionally also scribble a background mask where the color should remain unchanged. 
These masks do not need to be detailed; instead, a few coarse scribbles for each mask suffice. 
The user provides these inputs through a user interface, which we discuss in the appendix. 
Given the desired target color and foreground/background masks, we seek to update the neural network $\network$ and the latent color vector $\colorcode$ for the object instance to respect the user constraints. 

Let $\vc_f$ be the desired color for a ray $\ray$ at a pixel location within the foreground mask provided by the user scribble and let $y_f=\left\{(\ray,\vc_f)\right\}$ be the set of \newtext{ray color pairs} provided by the entire user scribble. 
Furthermore, for a ray $\ray$ at a pixel location in the background mask, let $\vc_b$ be the original rendered color at the ray location. 
Let $y_b=\left\{(\ray,\vc_b)\right\}$ be the set of rays and colors provided by the background user scribble. 

Given the user edit inputs \newtext{$\left(y_f, y_b\right)$}, we define our reconstruction loss as the sum of squared-Euclidean distances between the output colors from the compositing operation $\colorout$ to the target foreground and background colors:
\begin{align}
    \mathcal{L}_{\textrm{rec}} = \sum_{(\ray,\vc_f) \in y_f} \left|\left|\colorout\left(\ray, \shapecode, \colorcode\right) - \vc_f\right|\right|^2 \nonumber\\ 
    +  \sum_{(\ray,\vc_b) \in y_b} \left|\left|\colorout{\left(\ray, \shapecode, \colorcode\right)} - \vc_b\right|\right|^2.
        \label{eqn:color_edit}
\end{align}

Furthermore, we define a regularization term $\mathcal{L}_{\textrm{reg}}$ to discourage large deviations from the original model by penalizing the squared difference between original and updated model weights.

We define our {\em color editing loss} as the sum of our reconstruction loss and our regularization loss 
\begin{equation} \mathcal{L}_{\textrm{color}} = \mathcal{L}_{\textrm{rec}} + \lambda_{\textrm{reg}} \cdot \mathcal{L}_{\textrm{reg}}.
\end{equation}
We optimize this loss over the latent color vector $\colorcode$ and $\network_\text{rad}$ with $\lambda_{\textrm{reg}} = 10$.

\subsection{Shape Editing Loss}

For editing shapes, we describe two operations -- shape part removal and shape part addition, which we outline next.

\myparagraph{Shape part removal.}
To remove a shape part, the user scribbles over the desired removal region in a rendered view via the user interface. We take the scribbled regions of the view to be the foreground mask, and the non-scribbled regions of the view as the background mask. To construct the editing example, we whiten out the regions corresponding to the foreground mask.  

Given the editing example, we optimize a density-based loss that encourages the inferred densities to be sparse. 
Let $\sigma_\ray$ be a vector of inferred density values for sampled points along a ray $\ray$ at a pixel location and let $y_f$ be the foreground set of rays for the entire user scribble.

We define the density loss $\mathcal{L}_{\textrm{dens}}$ as the sum of entropies of the predicted density vectors $\sigma_{\ray}$ at foreground ray locations $\ray$,
\begin{equation}
    \mathcal{L}_{\textrm{dens}} = 
    -\sum_{\ray\in y_f} \sigma_\vr^\intercal \log{\left(\sigma_{\vr}\right)},
    \label{eqn:shape_edit}
\end{equation}
where we normalize all density vectors to be unit length. 
Penalizing the entropy along each ray encourages the inferred densities to be sparse, causing the model to predict zero density on the removed regions.

We define our {\em shape removal loss} as the sum of our reconstruction, density, and our regularization losses 
\begin{equation}\mathcal{L}_{\textrm{remove}} = \mathcal{L}_{\textrm{rec}} + \lambda_{\textrm{dens}} \cdot \mathcal{L}_{\textrm{dens}} + \lambda_{\textrm{reg}} \cdot \mathcal{L}_{\textrm{reg}}.
\end{equation}
We optimize this loss over $\network_{\text{dens}}$ and $\network_{\text{fuse}}$ with $\lambda_{\textrm{dens}} = 0.01$ and $\lambda_{\textrm{reg}} = 10$.

The above method of obtaining the editing example assumes that the desired object part to remove does not occlude any other object part. We describe an additional slower method for obtaining the editing example which deals with occlusions in the appendix.

\myparagraph{Shape part addition.}
To add a local part to a shape instance, we fit our network to a composite image comprising a region from a new object pasted into the original. To achieve this, the user first selects a original rendered view to edit. Our interface displays different instances under the same viewpoint and the user selects a new instance from which to copy. Then, the user copies a local region in the new instance by scribbling on the selected view. Finally, the user scribbles in the original view to select the desired paste location. For a ray in the paste location in the modified view, we render its color by using the shape code from the new instance and the color code from the original instance. We denote the modified regions of the composite view as the foreground region, and the unmodified regions as the background region.

We define our {\em shape addition loss} as the sum of our reconstruction and our regularization losses 
\begin{equation}
    \mathcal{L}_{\textrm{add}} = \mathcal{L}_{\textrm{rec}} +  \lambda_{\textrm{reg}} \cdot \mathcal{L}_{\textrm{reg}}
\end{equation}
and optimize over $\network_{\text{dens}}$ and $\network_{\text{fuse}}$ with $\lambda_{\textrm{reg}} = 10$.

\newnewtext{
We note that this shape addition method can be slow due to the large number of training iterations. In the appendix, we describe a faster but less effective method which encourages inferred densities to match the copied densities.}

Please refer to our \href{https://youtu.be/9qwRD4ejOpw}{video} to see our editing demo in action.

\section{Experiments}
\lblsec{exp}

In this section, we show the qualitative and quantitative results of our approach, perform model ablations, and compare our method to several baselines.

\myparagraph{\newtext{Datasets.}}
\newtext{We demonstrate our method on three publicly available datasets of varying complexity: chairs from the PhotoShape dataset~\cite{park2018photoshape} (large appearance variation), chairs from the Aubry chairs dataset~\cite{aubry2014seeing,dosovitskiy2015learning} (large shape variation), and cars from the GRAF CARLA dataset~\cite{schwarz2020graf,dosovitskiy2017carla} (single view per instance). For the PhotoShape dataset, we use $100$ instances with $40$ training views per instance. For the Aubry chairs dataset, we use $500$ instances with $36$ training views per instance. For the CARLA dataset, we use $1{,}000$ instances and have access to only a single training view per instance. For this dataset, to encourage color consistency across views, we regularize the view direction dependence of radiance, which we further study in the appendix. Furthermore, due to having access to only one view per instance, we forgo quantitative evaluation on the CARLA dataset and instead provide a qualitative evaluation.}

\myparagraph{Implementation details.}
Our shared shape network, instance-specific shape network, and fusion shape networks $\network_\text{share}, \network_\text{inst}, \network_\text{fuse}$ are all $4$ layers deep, $256$ channels wide MLPs with ReLU activations and outputs $256$ dimensional features. The shape and color codes are both $32$-dimensional and jointly optimized with the conditional radiance field model using the Adam optimizer~\cite{kingma2014adam} and a learning rate of $10^{-4}$. For each edit, we use Adam to optimize the parameters with a learning rate of $10^{-2}$. Additional implementation details are included in the appendix.

\subsection{Conditional \newtext{R}adiance \newtext{F}ield \newtext{T}raining}
\lblsec{multnerf}
\begin{table}[t]
    \centering 
    \resizebox{0.72\width}{!}
    {
    \begin{tabular}{l c c c c}
    \multicolumn{1}{c}{} &
    \multicolumn{2}{c}{PhotoShapes~\cite{park2018photoshape}} & 
    \multicolumn{2}{c}{Aubry \etal~\cite{aubry2014seeing}} \\
    \cmidrule(r){2-3}
    \cmidrule(r){4-5}
    &  PSNR $\uparrow$ & LPIPS $\downarrow$ &  PSNR $\uparrow$ & LPIPS $\downarrow$ \\ \hline 
    1) Single NeRF~\cite{mildenhall2020nerf}  & $17.81$ & $0.435$ & $14.26$ & $0.390$ \\ 
    2) + Learned Latent Codes  & $36.50$ & $0.029$  & $20.93$ & $0.164$ \\ 
    3) + Sep.\ Shape/Color Codes & $36.88$ & $0.028$ & $21.54$ & $0.153$ \\  
    4) + Share./Inst.\ Net (Ours) & $\textbf{37.67}$ & $\textbf{0.022}$ & $\textbf{21.78}$ & $\textbf{0.141}$ \\ \hline
    5) NeRF Separate Instances & $37.31$ & $0.035$ & $24.15$ & $0.041$ \\
    \vspace{-15pt}
    \end{tabular}
    }
    \caption{
    {\bf Ablation study.} \newtext{We evaluate our model and several ablations on view reconstruction.} Notice how separating the shape and color codes and using the shared/instance network improves the view synthesis quality. %
    Our model even outperforms single-instance NeRF models (each trained on one object).
    }
    \vspace{-12pt}
    \lbltbl{modelablation}
\end{table}

Our method accurately models the shape and appearance differences across instances. 
\newtext{To quantify this, we train our conditional radiance field on the PhotoShapes~\cite{park2018photoshape} and Aubry chairs~\cite{aubry2014seeing} datasets and evaluate the rendering accuracy on held-out views over each instance.} In \reftbl{modelablation}, we measure the rendering quality with two metrics: PSNR and LPIPS~\cite{zhang2018unreasonable}.
\newnewtext{In the appendix, we provide additional evaluation using the SSIM metric~\cite{wang2004image} in \reftbl{suppmodelablation} and visualize reconstruction results in Figures \ref{fig:supp-reconstructions12}-\ref{fig:supp-reconstructions910}.}  %
We find our model renders realistic views of each instance and, \newtext{on the PhotoShapes dataset,} matches the performance of training independent NeRF models for each instance.

We report an ablation study over the architectural choices of our method in \reftbl{modelablation}. \newtext{First, we train a standard NeRF~\cite{mildenhall2020nerf} over each dataset (Row 1).} Then, we add a $64$-dimensional learned code for each instance to the standard NeRF and jointly train the code and the NeRF (Row 2). The learned codes are injected wherever positional or directional embeddings are injected \newtext{in the original NeRF model}. While this choice is able to model the shape and appearance differences across the instances, we find that adding separate shape and color codes for each instance (Row 3) and further using a shared shape branch 
(Row 4) improves performance. 
Finally, we report performance when training independent NeRF models on each instance separately (Row 5). \newtext{In these experiments, we increase the width of the layers in the ablations to keep the number of parameters approximately equal across experiments.} 
Notice how our conditional radiance network outperforms all ablations.

\newtext{Moreover, we find that our method scales well to more training instances. When training with all $626$ instances of the PhotoShape dataset, our method achieves reconstruction PSNR $35.79$. We find that the shared shape branch helps our model scale to more instances. In contrast, a model trained without the shared shape branch achieves PSNR $33.91$.}

\subsection{Color Edits}
\lblsec{coloredits}
\begin{table}
    \centering 
    \resizebox{0.9\width}{!}
    {
    \begin{tabular}{l c c c}
            &  PSNR $\uparrow$ & LPIPS $\downarrow$  \\ \hline 
    Model Rewriting~\cite{bau2020rewriting} & $18.42$ & $0.325$ \\
    Finetuning Single-Instance NeRF & $29.53$ & $0.068$ \\
    Only Finetune Color Code & $26.29$ & $0.090$ \\
    Finetuning All Weights & $31.00$ & $0.050$  \\
    Our Method  & $\textbf{35.25}$ & $\textbf{0.027}$ \\
    \end{tabular}
    }
    \vspace{-5pt}
    \caption{
    {\bf Color editing quantitative results.} 
    We evaluate color editing of a source object instance to match a target instance. Our method outperforms the baselines on all criteria. }
    \lbltbl{coloredits}
    \vspace{-12pt}
\end{table}

\begin{figure*}
  \centering
     \vspace{-20pt}
   \includegraphics[width=\linewidth]{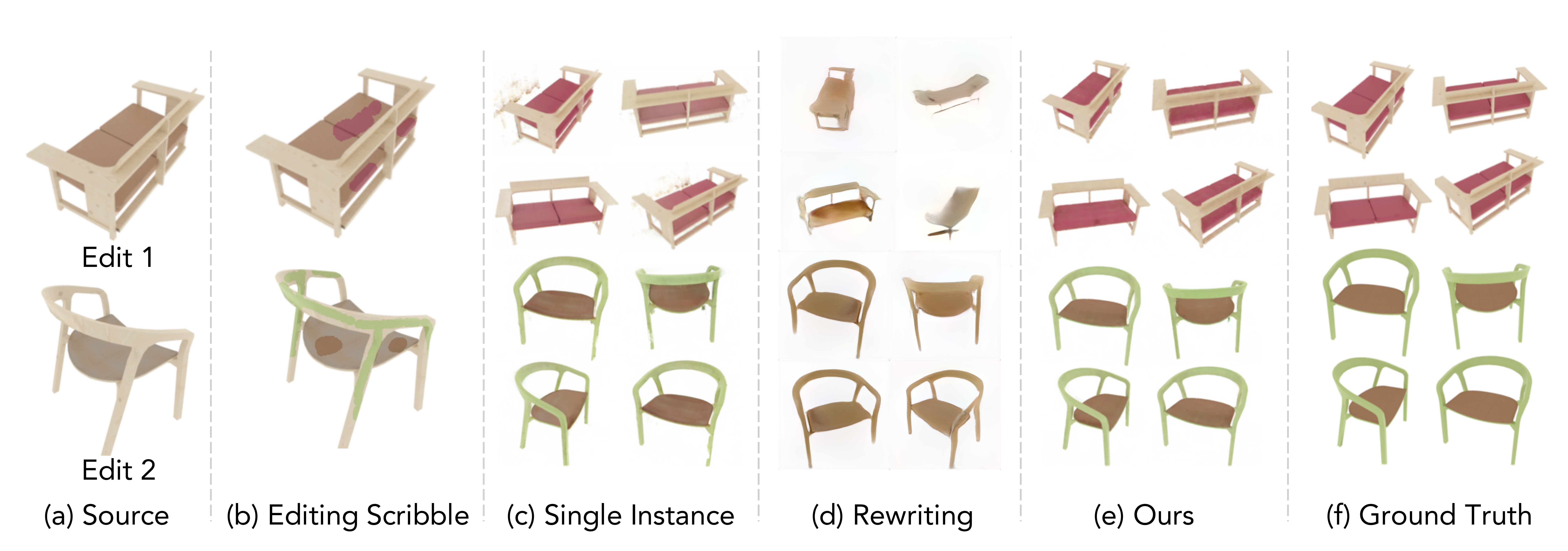}
   \vspace{-20pt}
  \caption{
  {\bf Color editing qualitative results.} 
  We visualize color editing results where the goal is to match a source instance's colors to a target. Our method accurately captures the colors of the target instance given scribbles over one view. Notice how (c) Editing a Single-Instance NeRF causes visual artifacts, and (d) Rewriting a GAN~\cite{bau2020rewriting} fails to propagate the edit to unseen views and generates unrealistic outputs.
  }
  \lblfig{coloredits}
\end{figure*}

Our method both propagates edits to the desired regions of the instance and generalizes to unseen views of the instance. We show several example color edits in \reffig{coloredits}. 
To evaluate \newtext{our} choice of optimization parameters, we conduct an ablation study to \newtext{quantify} our edit quality. 

For a source \newtext{PhotoShapes} training instance, we first find an \emph{unseen} target instance in the \newtext{PhotoShapes} chair dataset with an identical shape but a different color. Our goal is to edit the source training instance to match the target instance across all viewpoints. We conduct three edits and show visual results on two: changing the color of a seat from brown to red (Edit 1), and darkening the seat and turning the back green (Edit 2). The details and results of the last edit can be found in the appendix. After each edit, we render $40$ views from the ground truth instance and the edited model, and \newtext{quantify the} difference. The averaged results over the three edits are summarized in \reftbl{coloredits}.

We find that finetuning only the color code is unable to fit the desired edit. On the other hand, changing the entire network leads to large changes in the shape of the instance, as finetuning the earlier layers of the network can affect the downstream \newtext{density} output. 

Next, we compare our method against two baseline methods: editing a single-instance NeRF and editing a GAN.

\myparagraph{Single-instance NeRF baseline.}
\newtext{W}e train a NeRF to model the source instance we would like to edit, and then apply our editing method to the single instance NeRF. The single instance NeRF shares the same architecture as our model. 

\myparagraph{GAN editing baselines.} \newtext{We also compare our method to the 2D GAN-based editing method based on Model Rewriting~\cite{bau2020rewriting}. We first train a StyleGAN2 model~\cite{Karras2019stylegan2} on the images of the PhotoShapes dataset~\cite{park2018photoshape}. Then, we project unedited test views of the source instance into latent and noise vectors, using the StyleGAN2 projection method~\cite{Karras2019stylegan2}. Next, we invert the source and target view into its latent and noise vectors. With these image/latent pairs, we follow the method of Bau \etal~\cite{bau2020rewriting} and optimize the network to paste the regions of the target view onto the source view. After the optimization is complete, we feed the test set latent and noise vectors into the edited model to obtain edited views of our instance. In the appendix, we provide an additional comparison against naive finetuning of the whole generator.}

These results \newtext{are} visualized in \reffig{coloredits} and in \reftbl{coloredits}.  A single-instance NeRF is unable to find an change in the model that generalizes to other views, due to the lack of category-specific appearance prior. Finetuning the model can lead to artifacts in other views of the model and can lead to color inconsistencies across views. Furthermore, 2D GAN-based editing methods fail to correctly modify the color of the object or maintain shape consistency across views, due to the lack of 3D representation.

\subsection{Shape Edits}
\lblsec{shapeedits}
\begin{table}
    \centering 
    \vspace{-12pt}
    \resizebox{0.85\width}{!}
    {
    \begin{tabular}{l c c c c }
            &  PSNR $\uparrow$ & LPIPS $\downarrow$ & Time (s) $\downarrow$ \\ \hline 
    Only Finetune Shape Code & $22.07$ & $0.119$ & $36.9$  \\
    Only Finetune $\network_\text{dens}$ &  $21.84$& $0.118$ & $\textbf{27.2}$ \\
    Finetuning All Weights & $20.31$ & $0.117$ & $66.4$  \\
    Our Method & $\textbf{24.57}$ & $\textbf{0.081}$ & $37.4$  \\
    \end{tabular}
    }
    \vspace{-5pt}
    \caption{\newtext{
    {\bf Shape editing quantitative results.} 
    Notice how our hybrid network update approach achieves high visual edit quality while balancing computational cost.
    }
    }
    \lbltbl{shapeedits}
    \vspace{-10pt}
\end{table}

\begin{figure*}
  \centering
\includegraphics[width=\linewidth]{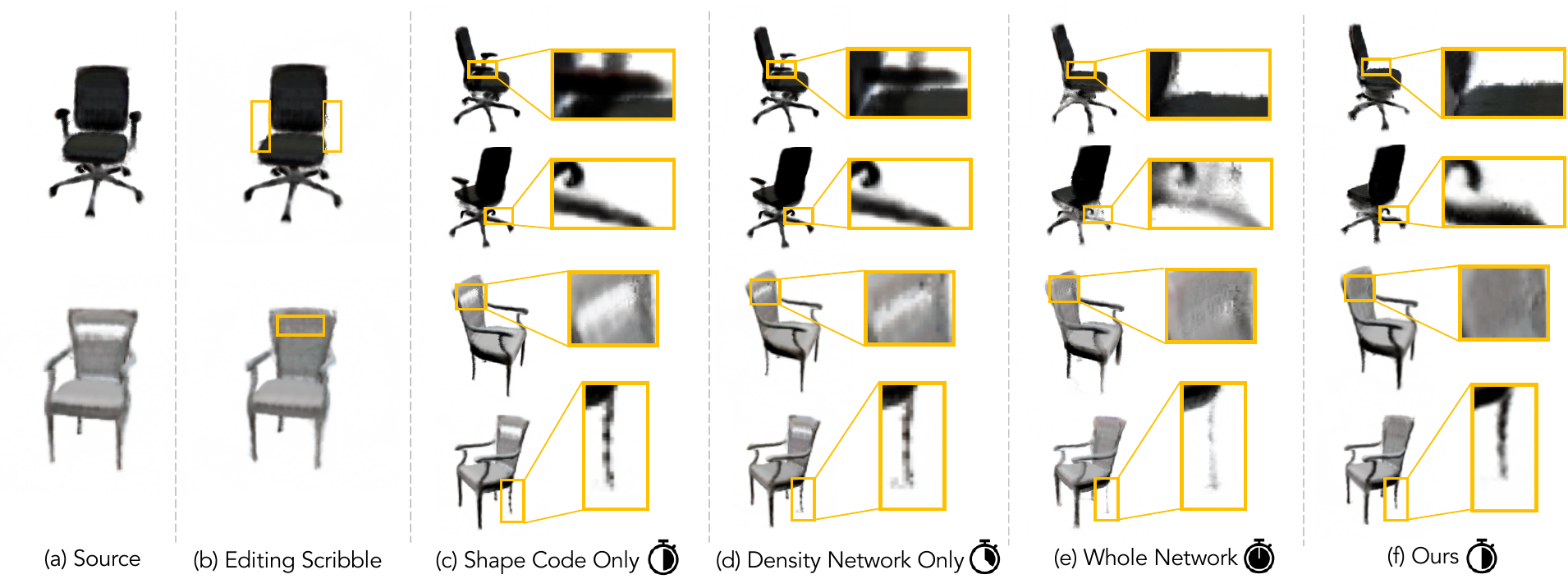}
  \captionof{figure}{
  {\bf Shape editing qualitative results.} 
  Our method successfully removes the arms and fills in the hole of a chair. Notice how only optimizing the shape code or branch are unable to fit both edits. Optimizing the whole network is slow and causes unwanted changes in the instance.
  }
  \vspace{-5pt}
  \lblfig{shapeedits}
\end{figure*}

\begin{figure*}[h]
  \centering
  \includegraphics[width=\linewidth]{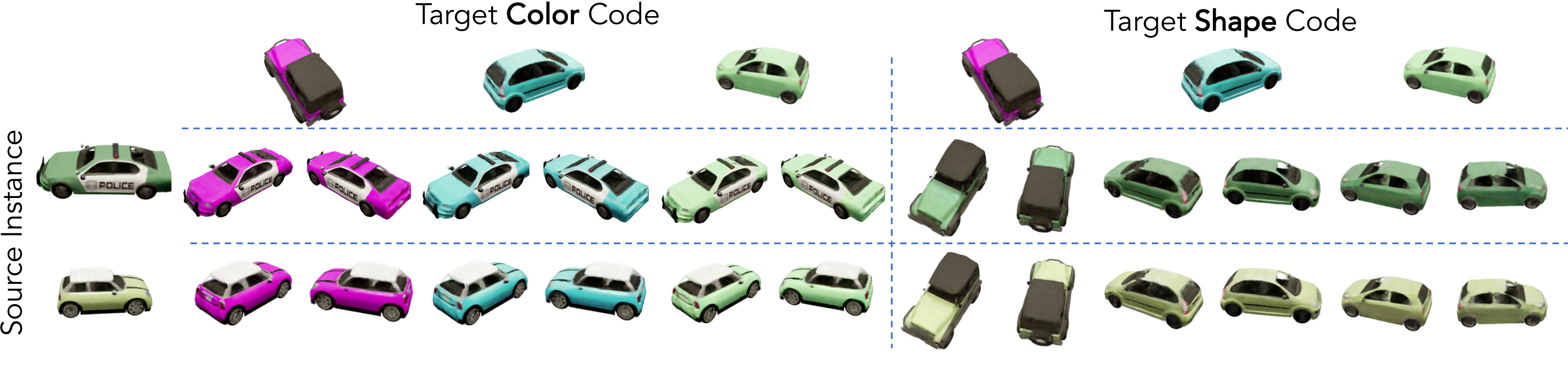}
  \vspace{-15pt}
  \caption{{\bf Shape and color transfer results.} Our model transfers the shape and color from target instances to a given source instance. When a source's color code is swapped with a target's, the shape remains unchanged, and vice versa.}
  \lblfig{disentangle}
  \vspace{-7pt}
\end{figure*}

\begin{figure}[h]
  \centering
  \includegraphics[width=\linewidth]{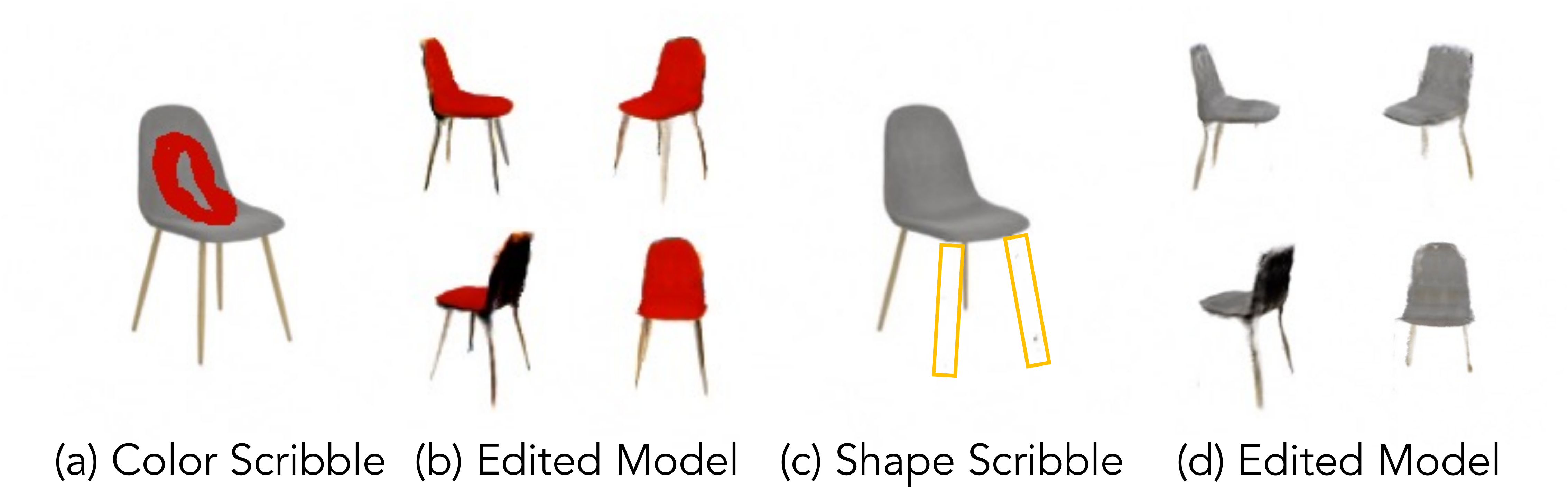}
  \caption{
  \newtext{{\bf Real image editing results.}  Our method first finetunes a conditional radiance field to match a real still image input. Editing the resulting radiance field successfully changes the chair seat color to red and removes two of the chair's legs.}
  }
  \lblfig{realimages}
  \vspace{-15pt}
\end{figure}

Our method is also able to learn to edit the shape of an instance and propagate the edit to unseen views.  \newtext{We show several shape editing examples in \reffig{shapeedits}.}
\newtext{Similar to our analysis of color edits, we evaluate our choice of weights to optimize. For a source Aubry chair dataset training instance, we find an unseen target instance with a similar shape. We then conduct an edit to change the shape of the source instance to the target instance, and quantify the difference between the rendered and ground truth views. The averaged results across three edits are summarized in \reftbl{shapeedits} and results of one edit are visualized in the top of \reffig{shapeedits}.}

We find that the approaches of only optimizing the shape code and only optimizing $\network_\text{dens}$ are unable to fit the desired edit, and instead leave the chair mostly unchanged. \newtext{Optimizing the whole network leads to removal of the object part, but causes unwanted artifacts in the rest of the object.} Instead, our method correctly removes the arms and fills the hole of the chairs, and generalizes this edit to unseen views of each instance.

\subsection{Shape/Color Code Swapping}
\lblsec{disentangle}
Our model succeeds in disentangling shape and color. When we change the color code input to the conditional radiance field while keeping the shape code unchanged, the resulting rendered views remain consistent in shape. Our model architecture enforces this consistency, as the density output that governs the shape of the instance is independent of the color code. 

When changing the shape code input of the conditional radiance field while keeping the color code unchanged, the rendered views remain consistent in color. This is surprising because in our architecture, the radiance of a point is a function of both the shape code and the color code. Instead, the model has learned to disentangle color from shape when predicting radiance. 
These properties let us freely swap around shape and color codes, allowing for the transfer of shape and appearance across instances; we visualize this in \reffig{disentangle}. 

\subsection{\newtext{Real Image Editing}}
\newtext{
We demonstrate how to infer and edit extrapolated novel views for a single real image given a trained conditional radiance field. 
We assume that the single image has attributes similar to the conditional radiance field's training data (\eg object class, background). First, we estimate the image's viewpoint by manually selecting a training set image with similar object pose. In practice, we find that a perfect pose estimation is not required.
With the posed input image, we finetune the conditional radiance field by optimizing the standard NeRF photometric loss with respect to the image. When conducting this optimization, we first optimize the shape and color codes of the model, while keeping the MLP weights fixed, and then optimize all the parameters jointly. This optimization is more stable than the alternative of optimizing all parameters jointly from the start.
Given the finetuned radiance field, we proceed with our editing methods to edit the shape and color of the instance. We demonstrate our results of editing a real photograph in \reffig{realimages}. 
}

\section{Discussion}
\lblsec{discussion}
We have introduced an approach for learning conditional radiance fields from a collection of 3D objects. 
Furthermore, we have shown how to perform intuitive %
editing operations using our learned disentangled representation. 
\newtext{
One limitation of our method is the interactivity of shape editing. Currently, it takes over a minute for a user to get feedback on their shape edit. The bulk of the editing operation computation is spent on rendering views, rather than editing itself. We are optimistic that NeRF rendering time improvements will help~\cite{neff2021donerf,yu2021plenoctrees}. Another limitation is our method fails to reconstruct novel object instances that are very different from other class instances.}
\newtext{Despite these limitations,} our approach opens up new avenues for exploring other advanced editing operations, such as relighting and changing an object's physical properties for animation.

\myparagraph{Acknowledgments.} Part of this work while SL was an intern at Adobe Research. We would like to thank William T. Freeman for helpful discussions.

\newpage

{\small
\bibliographystyle{ieee_fullname}
\bibliography{main}
}

\newpage
\newpage
\appendix
In this appendix, we provide the following:

\begin{itemize}
    \item \newtext{Additional experimental details on datasets, model training, and model editing (\refsec{additionaldetails}).}
    \item Additional evaluations of our method and additional ablations. (\refsec{additionaleval}).
    \item \newtext{Additional methods for shape editing. (\refsec{additionalmethod}).}
    \item A description of our user interface. (\refsec{ui}).
    \item Additional visualizations of color edits with our approach (\refsec{additionalcolor}). 
    \item \newtext{Additional visualizations of shape edits with our approach (\refsec{additionalshape}).}
    \item Additional visualizations of color and shape swapping edits with our approach (\refsec{additionalswap}). 
    \item A visualization of reconstructions with our conditional radiance field (\refsec{reconstructions}). 
    \item Changelog (\refsec{changelog})
\end{itemize}

\section{Additional Experimental Details}
\lblsec{additionaldetails}
\myparagraph{Dataset rendering details.}
For the PhotoShape dataset~\cite{park2018photoshape}, we use Blender~\cite{blendertutorial} to render $40$ views for each instance with a clean background. To obtain the clean background, we obtain the occupancy mask and set everywhere else to white. For the Aubry chairs dataset~\cite{aubry2014seeing}, we resize the images from $600\times600$ resolution to $400\times400$ resolution and take a $256\times256$ center crop. For the CARLA~\cite{dosovitskiy2017carla} dataset, we train our radiance field on exactly the same dataset as GRAF~\cite{schwarz2020graf}.

\myparagraph{Conditional radiance field training details.}
As in Mildenhall \etal~\cite{mildenhall2020nerf}, we train two networks, a coarse network $\mathcal{F}_{\textrm{coarse}}$ to estimate the density along the ray, and a fine network $\mathcal{F}_{\textrm{fine}}$ that renders the rays at test time. We use stratified sampling to sample points for the coarse network and sample a hierarchical volume using the coarse network's density outputs. 
The rendered outputs of these networks for an input ray $\vr$ are given by $\hat{C}_{\textrm{coarse}}(\vr)$ and $\hat{C}_{\textrm{fine}}(\vr)$, respectively. 
The networks are jointly trained with the shape and color codes to optimize a photometric loss. During each training iteration, we first sample an object instance $k \in \{1, \dots, K \}$ and obtain the corresponding shape code $\shapecode_k$ and color code $\colorcode_k$. 
Then, we sample a batch of training rays from the set of all rays $R_k$ belonging to the instance $k$, and optimize both networks using a photometric loss, which is the sum of squared-Euclidean distances between the predicted colors and ground truth colors,
\begin{align}
    \mathcal{L_{\textrm{train}}} = \sum_{k=1}^{K} \biggl[ \: \sum_{\vr \in R_k} ||\hat{C}_{\textrm{coarse}}(\vr, \shapecode_k, \colorcode_k) - C(\vr)||_2^2
    \nonumber\\ 
    + ||\hat{C}_{\textrm{fine}}(\vr, \shapecode_k, \colorcode_k) - C(\vr)||_2^2 \biggr].
    \lbleq{training}
\end{align}

\newtext{
When training all radiance field models, we optimize our parameters using Adam~\cite{kingma2014adam} with a learning rate of $10^{-4}$, $\beta_1 = 0.9$, $\beta_2 = 0.999$, and $\eps = 10^{-8}$. We train our models until convergence, which on average is around $1$M iterations.}

\myparagraph{Conditional radiance field editing details.}
During editing, we keep the coarse network fixed and edit the fine network $\mathcal{F}_{\textrm{fine}}$ only. We increase the learning rate to $10^{-2}$, and we optimize network components and codes for $100$ iterations, keeping all other hyperparameters the same. To obtain training rays, we first randomly select an edited view of the instance, then randomly sample batches of rays from the subsampled set of foreground and background rays. 

\myparagraph{Conditional radiance field architecture details.}
\newtext{
Like the original NeRF paper~\cite{mildenhall2020nerf}, we use skip connections in our architecture: the shape code and embedded input points are fed into the fusion shape network as well as the very beginning of the network. 
}

\newtext{
Furthermore, in our model architecture, we introduce a bottleneck that allows feature caching to be computationally feasible. The input to the color branch is an $8$-dimensional vector, which we cache during color editing. 
}

\newtext{
Last, when injecting a shape or color code to a layer, we run the code through a linear layer with ReLU nonlinearity and concatenate the output with the input to the layer.
}

\myparagraph{GAN editing details.}
In our GAN experiments, we use the default StyleGAN2~\cite{Karras2019stylegan2} configuration on the PhotoShapes dataset and train for $300{,}000$ iterations. For model rewriting~\cite{bau2020rewriting}, we optimize the $8$-th layer for $2{,}000$ iterations with a learning rate of $10^{-2}$.

\myparagraph{View direction dependence.}
\begin{figure*}
  \centering
   \includegraphics[width=\linewidth]{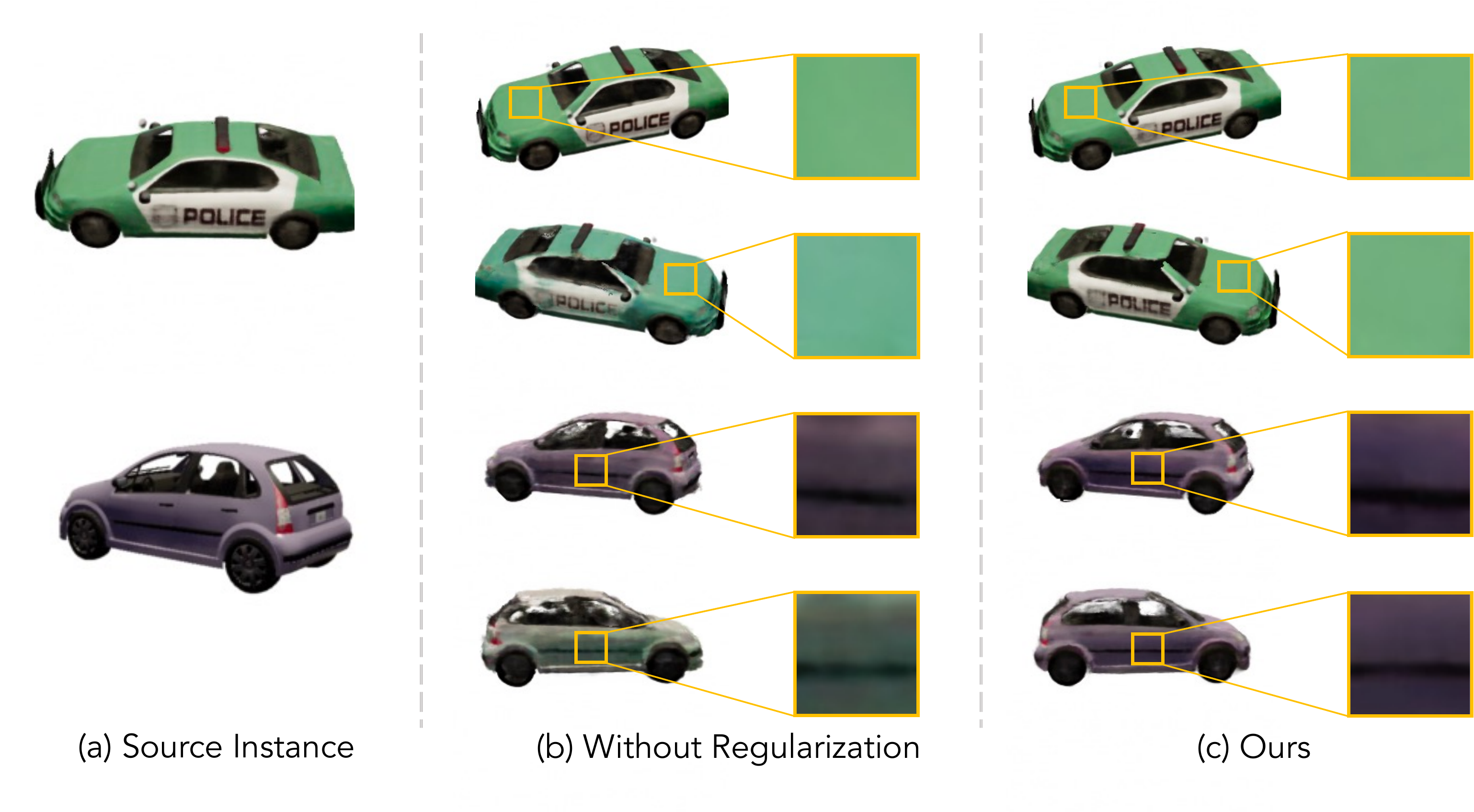}
  \caption{\newtext{
  {\bf CARLA~\cite{dosovitskiy2017carla} dataset radiance view dependence regularization.} 
  We show synthesized views from an unregularized conditional radiance field and a regularized conditional radiance field trained on one view per instance. Notice how the regularized model is consistent in color across views while the unregularized model is not, hallucinating between green and blue (top) and purple and green (bottom). 
  }}
  \lblfig{supp-viewdirreg}
\end{figure*}

On the CARLA dataset~\cite{dosovitskiy2017carla, schwarz2020graf}, we find that having only one training view per instance can cause color inconsistency across rendered views. To address this, we regularize the view dependence of the radiance $\vc$ with an additional self-consistency loss that encourages the model to predict for a point $\vx$ similar radiance across viewing directions $\vd$. This loss penalizes, for point $\vx$ and viewing direction $\vd$ in the radiance field, the squared difference between the sampled radiance $\vc(\vx, \vd)$ and the average radiance of the point $\vx$ over all viewing directions. Specifically, given radiance field inputs $\vx, \vd$, we minimize 
\begin{equation*}
    \mathbb{E}_{\vx \sim p_\vx; \vd \sim p_\vd} \left[ \left\Vert \vc(\vx, \vd) - \mathbb{E}_{\vd' \sim p_\vd}[\vc(\vx, \vd') \right\Vert^2 \right]
\end{equation*} where $p_\vx$ is the probability distribution over points $\vx$ and $p_\vd$ is the probability distribution over all viewing directions $\vd$. During training, we approximate this loss by first sampling a training point $\vx$ and viewing direction $\vd$, then approximating the inner expectation $\mathbb{E}_{\vd' \sim p_\vd}[\vc(\vx, \vd')]$ by sampling $K$ viewing directions $\vd_i \sim p_\vd$, and taking $$\mathbb{E}_{\vd' \sim p_\vd}[\vc(\vx, \vd')] \approx  \frac{1}{K}\sum_{i=1}^{K}\vc(\vx, \vd_i).$$ In our experiments, we use $K = 64$. We visualize results with and without this regularization in \reffig{supp-viewdirreg}.

\section{Additional Evaluations}
\lblsec{additionaleval}
\begin{table*}[t]
    \centering 
    \resizebox{\width}{!}
    {
    \begin{tabular}{lrrrrrr}
    \multicolumn{1}{c}{} &
    \multicolumn{3}{c}{PhotoShapes~\cite{park2018photoshape}} & 
    \multicolumn{3}{c}{Aubry \etal~\cite{aubry2014seeing}} \\ 
    \cmidrule(r){2-4}
    \cmidrule(r){5-7}
            &  PSNR $\uparrow$ & SSIM $\uparrow$ & LPIPS $\downarrow$ &  PSNR $\uparrow$ & SSIM $\uparrow$ & LPIPS $\downarrow$ \\ \hline 
    1) Single NeRF~\cite{mildenhall2020nerf}  & $17.81$ & $0.836$ & $0.435$ & $14.26$ & $0.814$ & $0.390$ \\ 
    2) + Learned Latent Codes  & $36.50$ & $0.979$ & $0.029$  & $20.93$ & $0.892$ & $0.164$ \\ 
    3) + Sep.\ Shape/Color Codes & $36.88$ & $0.980$ & $0.028$ & $21.54$ & $0.898$ & $0.144$ \\  
    4) + Shar./Inst.\ Net (Ours) & $\textbf{37.67}$ & $\textbf{0.982}$ & $\textbf{0.022}$ & $\textbf{21.78}$ & $\textbf{0.900}$ & $\textbf{0.141}$ \\ \hline
    5) NeRF Separate Instances & $37.31$ & $0.972$ & $0.035$ & $24.15$ & $0.963$ & $0.041$ \\
    \vspace{-15pt}
    \end{tabular}
    }
    \caption{
    {\bf Conditional radiance field ablation study.} We evaluate our model and several ablations on novel view synthesis. Notice how separating the shape and color codes and using the shared/instance network improves the view synthesis quality. %
    }
    \vspace{-12pt}
    \lbltbl{suppmodelablation}
\end{table*}

\begin{table}
    \centering 
    \resizebox{0.8\width}{!}
    {
    \begin{tabular}{l c c c}
            &  PSNR $\uparrow$ & SSIM $\uparrow$ & LPIPS $\downarrow$  \\ \hline 
    GAN-Finetuning & $19.64$ & $0.704$ & $0.255$  \\
    Model Rewriting~\cite{bau2020rewriting} & $18.42$ & $0.622$ & $0.325$ \\
    Finetuning Single-Instance NeRF & $29.53$ & $0.955$ & $0.068$ \\
    Only Finetune Color Code & $26.29$ & $0.968$ & $0.090$ \\
    Finetuning All Weights & $31.00$ & $0.957$ & $0.050$  \\
    Our Method  & $\textbf{35.25}$ & $\textbf{0.977}$ & $\textbf{0.027}$ \\
    \end{tabular}
    }
    \vspace{-5pt}
    \caption{
    {\bf Color editing quantitative results.} 
    We evaluate color editing of a source object instance to match a target instance. Our method outperforms the baselines on all criteria. }
    \lbltbl{suppcoloredits}
    \vspace{-12pt}
\end{table}

\begin{table}
    \centering 
    \resizebox{0.75\width}{!}
    {
    \begin{tabular}{l c c c c }
            &  PSNR $\uparrow$ & SSIM $\uparrow$ & LPIPS $\downarrow$ & Time (s) $\downarrow$ \\ \hline 
    Only Finetune Shape Code & $22.08$ & $0.931$ & $0.119$ & $36.9$  \\
    Only Finetune $\mathcal{F}_\text{dens}$ &  $21.84$ & $0.921$ & $0.118$ & $\textbf{27.2}$ \\
    Finetuning All Weights & $20.31$ & $0.910$ & $0.117$ & $66.4$  \\
    Our Method & $\textbf{24.57}$ & $\textbf{0.944}$ & $\textbf{0.081}$ & $37.4$  \\
    \end{tabular}
    }
    \vspace{-5pt}
    \caption{{\bf Shape editing quantitative results.} 
    Notice how our hybrid network update approach achieves high visual edit quality while balancing computational cost.
    }
    \lbltbl{suppshapeedits}
    \vspace{-12pt}
\end{table}

\myparagraph{SSIM metric.}
We report additional evaluation on model ablation, color editing, and shape editing results using SSIM~\cite{wang2004image}. Quantitative results can be found in this appendix's \reftbl{suppmodelablation} (model ablation), \reftbl{suppcoloredits} (color edits), and \reftbl{suppshapeedits} (shape edits). 

\myparagraph{Subsampling user constraints.}
During editing, we do not train on the whole foreground and background regions provided by the user, which can potentially decrease the quality of our edits. This is because training on fewer rays can cause the edit to propagate onto unwanted areas. For example, regions which the user specify as background, but are not in the set of sampled rays, can potentially be changed. However, we find that upon adding this optimization, the average PSNR over the three color edits decreases to $34.49$. 

\myparagraph{Additional GAN editing baselines.}
We compare our editing method against a naive generator fine-tuning method ~\cite{bau2020rewriting}. The method is identical to the model rewriting method, except instead of conducting a low-rank update of the weights of a particular layer, we freely optimize all the weights of the generator. This optimization is done over $10{,}000$ steps with a learning rate of $10^{-3}$. We report our results in \reftbl{suppcoloredits}. 

\section{\newtext{Additional Shape Editing Methods}}
\lblsec{additionalmethod}

\myparagraph{Shape removal.} \newnewtext{For shape removal method described in the main paper, we assume that there is nothing behind an object part that a user scribbles over, allowing us to replace the object part with a white background. However, in practice, a user may wish to remove an object part that is in front of another one. To handle such occlusion, we propose a separate procedure: for each ray in the foreground mask, we zero out the first mode of density along the ray. We define the first mode of density to start at the first point with nonzero-density up to the first subsequent point with zero density. We find that this procedure is effective but can be slow and may leave artifacts of incomplete removal.}

\myparagraph{Shape addition.}
\newnewtext{
For shape addition, our method for reconstructing a composite image leads to effective but slow edits. We propose an additional density-based loss which is faster but less effective in executing the edit. The method for obtaining the editing example is the same, but we now optimize a loss that encourages the density values in the modified regions of the composite view to match with the density values of the regions copied from. 
}

\newnewtext{
Specifically, let $y_f = \{ (\vr, \sigma_f) \}$ be the set of rays and densities in the foreground mask and $y_b = \{ (\vr, \sigma_b) \}$ be the set of rays and densities in the background mask. Here, $\sigma_f$ are density values of the rays copied from the new object instance, and $\sigma_b$ represent density values of the rays from the original instance. Furthermore, let $\sigma_\vr$ be the densities predicted by our model for ray $\vr$.  Again, densities are normalized to sum to one. 
}

\newnewtext{
We optimize a cross-entropy loss:
\begin{equation}
\mathcal{L}_{\textrm{dens}} = \sum_{(\vr, \sigma_f) \in y_f} - \sigma_f^T \log \sigma_\vr +  \sum_{(\vr, \sigma_b) \in y_b} -\sigma_b^T \log \sigma_\vr,
\end{equation}
which encourages the predicted densities to match the target densities for the edited regions and be unchanged for the unedited regions. 
}

\section{User Interface}
\lblsec{ui}

For our user interface, the user first picks an object instance they would like to edit. Our UI then displays several rendered views of that instance, and the user picks one view to edit. The user can then edit the selected view on an editing panel.%

We provide four types of user edits: color edits, shape removal, shape addition, and color/shape transfer. Next, we describe the user interactions for each type of edit.

\myparagraph{Color edits.} The user chooses the target color from a color palette. Then, the user specifies a foreground mask over the view by clicking the \texttt{edit color} button, selecting a brush color, and scribbling over parts of the object. 
Last, the user specifies the background mask by clicking the \texttt{BG} brush and scribbling over where they would like to keep the part unchanged. 

\myparagraph{Shape removal.} The user clicks the \texttt{remove shape} button and scribbles over parts of the image they would like to remove. %

\myparagraph{Shape addition.} The user clicks the \texttt{add shape} button and several instances to copy shape from will pop up. The user specifies a target instance they would like to copy from, and a view of that instance is shown. Then, the user scribbles over the object part they would like to copy, and clicks on the location of the source instance where they would like to paste.%

\myparagraph{Shape/Color transfer.} The user clicks either the \texttt{transfer color} button or the \texttt{transfer shape} button and several instances to transfer color/shape from will pop up. Then, the user clicks a desired target instance to transfer color/shape information. 

Once the user editing is done, the user will click the \texttt{execute} button to execute the desired edit. Our algorithm will then finetune the latent variables and network weights and update the renderings of the edited object. Please see our video \href{https://www.youtube.com/watch?v=9qwRD4ejOpw}{demo} for more details.

\section{Additional Color Edits}
\lblsec{additionalcolor}

\begin{figure*}
  \centering
   \includegraphics[width=\linewidth]{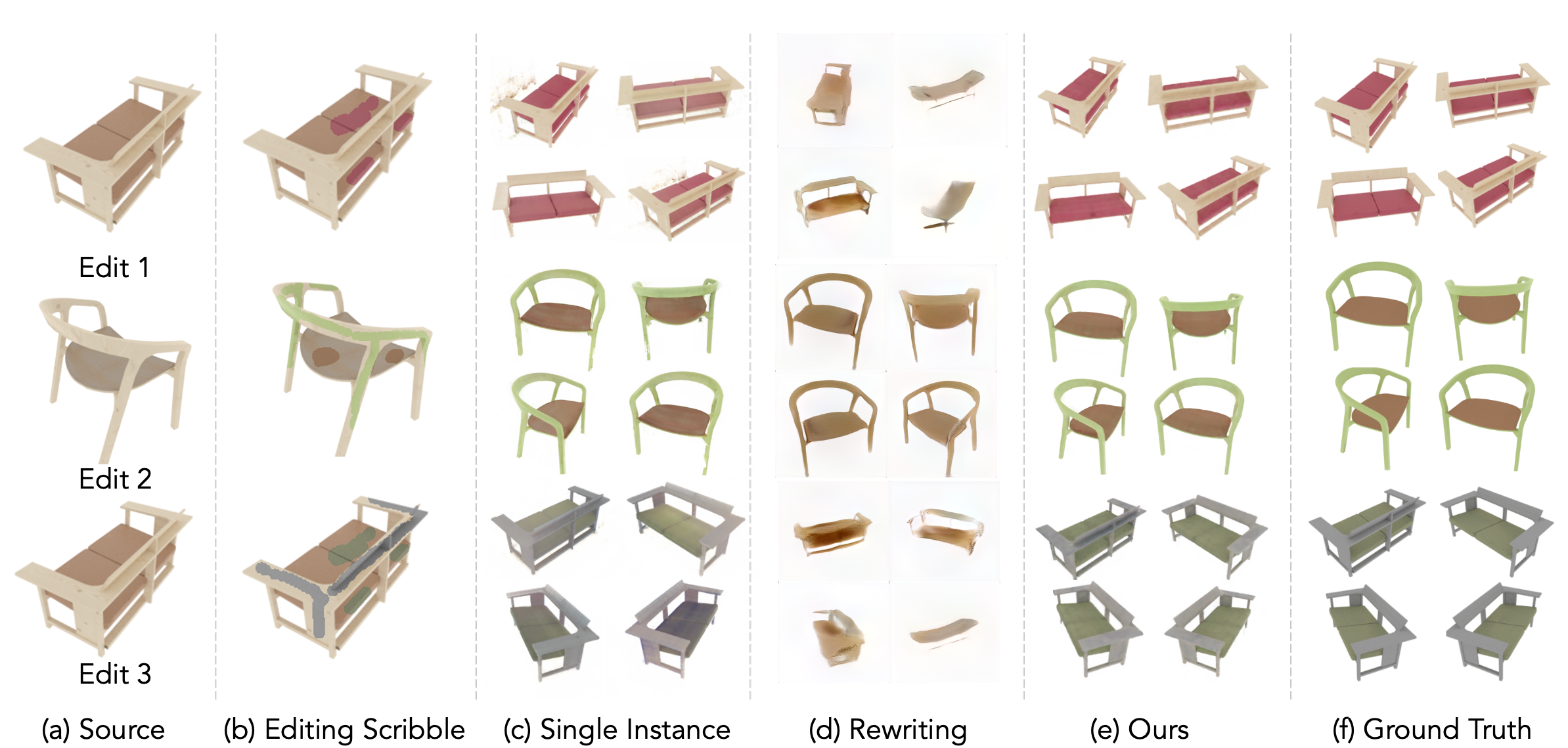}
  \caption{
  {\bf Color editing qualitative results.} 
  We visualize color editing results where the goal is to match a source instance's colors to a target. Our method accurately captures the colors of the target instance given scribbles over one view. Notice how (d) Rewriting a GAN~\cite{bau2020rewriting} fails to propagate the edit to unseen views and results in unrealistic generated outputs. Moreover, editing a single-instance NeRF causes visual floating artifacts (Edit 1) and non-transferring colors (Edit 3).
  }
  \lblfig{supp-figcoloredits}
\end{figure*}

\begin{table*}
    \centering 
    \resizebox{\textwidth}{!}
    {
    \begin{tabular}{lrrrrrrrrr}
    \toprule
    \multicolumn{1}{c}{} &
    \multicolumn{3}{c}{Edit 1} & 
    \multicolumn{3}{c}{Edit 2} & 
    \multicolumn{3}{c}{Edit 3} \\ %
    \cmidrule(r){2-4}
    \cmidrule(r){5-7}
    \cmidrule(r){8-10}
            & \multicolumn{1}{c}{PSNR}  &  \multicolumn{1}{c}{SSIM}  & \multicolumn{1}{c}{LPIPS} 
            & \multicolumn{1}{c}{PSNR}  &  \multicolumn{1}{c}{SSIM}  & \multicolumn{1}{c}{LPIPS} 
            & \multicolumn{1}{c}{PSNR}  &  \multicolumn{1}{c}{SSIM}  & \multicolumn{1}{c}{LPIPS} \\
    \midrule
    GAN-Finetuning & $21.42$  & $0.744$ & $0.218$ & $20.45$ & $0.730$ & $0.264$ & $18.91$ & $0.686$ & $0.251$ \\ %
    Model Rewriting & $20.44$ & $0.663$ & $0.301$ & $19.11$ & $0.659$ & $0.322$ & $18.03$ & $0.612$ & $0.315$ \\
    Finetuning Single-Instance NeRF & $28.22$ & $0.933$ & $0.125$ & $28.32$ & $0.956$ & $0.057$ & $29.88$ & $0.964$ & $0.051$ \\ %
    Only Finetune Color Code & $28.15$ & $0.975$ & $0.054$ & $27.77$ & $0.977$ & $0.097$ & $22.95$ & $0.953$ & $0.120$\\ %
    Finetuning All Weights & $33.36$ & $0.970$ & $0.029$ & $32.47$ & $0.967$ & $0.036$ & $27.17$ & $0.936$ & $0.086$ \\
    Our Method  & $\textbf{35.32}$ & $\textbf{0.978}$ & $\textbf{0.024}$ & $\textbf{35.72}$ & $\textbf{0.979}$ & $\textbf{0.027}$ & $\textbf{34.72}$ & $\textbf{0.975}$ & $\textbf{0.031}$  \\ %
    \bottomrule
    \end{tabular}
    }
    \caption{
    {\bf Color editing quantitative results.} 
    We evaluate color editing of a source object instance to match a target instance. Please refer to \reffig{supp-figcoloredits} (this supplemental) for a visualization of each of the edits. Notice that our method outperforms the baselines for all color edits on all criteria. }
    \lbltbl{supp-tblcoloredits}
\end{table*}

\myparagraph{Quantitative color editing evaluation.}
In this section, we provide the visualizations of all three color edits used for evaluation in the main paper. 
Visually, we again see that editing a single-instance NeRF leads to visual artifacts and visual inconsistencies across views. Similarly, GAN-based methods are unable to learn an edit that generalizes across views, likely due to their lack of a 3D representation. 
We visualize the results of the three edits in \reffig{supp-figcoloredits} and quantify them in \reftbl{supp-tblcoloredits}. 

In \reffig{supp-figcoloredits}, the first two rows visualize Edits 1 and 2 discussed in the main paper, and are identical to the visualizations in the main paper's \reffig{coloredits}. The last row of \reffig{supp-figcoloredits} visualizes Edit 3, which changes the seat of a chair from brown to green, then the chair back from beige to grey. 

In \reftbl{supp-tblcoloredits}, we quantify the quality of each of the four edits. The first two main columns correspond to the Edits 1 and 2 discussed in the main paper, while the last two main columns correspond to Edits 3 discussed in \refsec{additionalcolor}.

\begin{figure*}
  \centering
   \includegraphics[width=\linewidth]{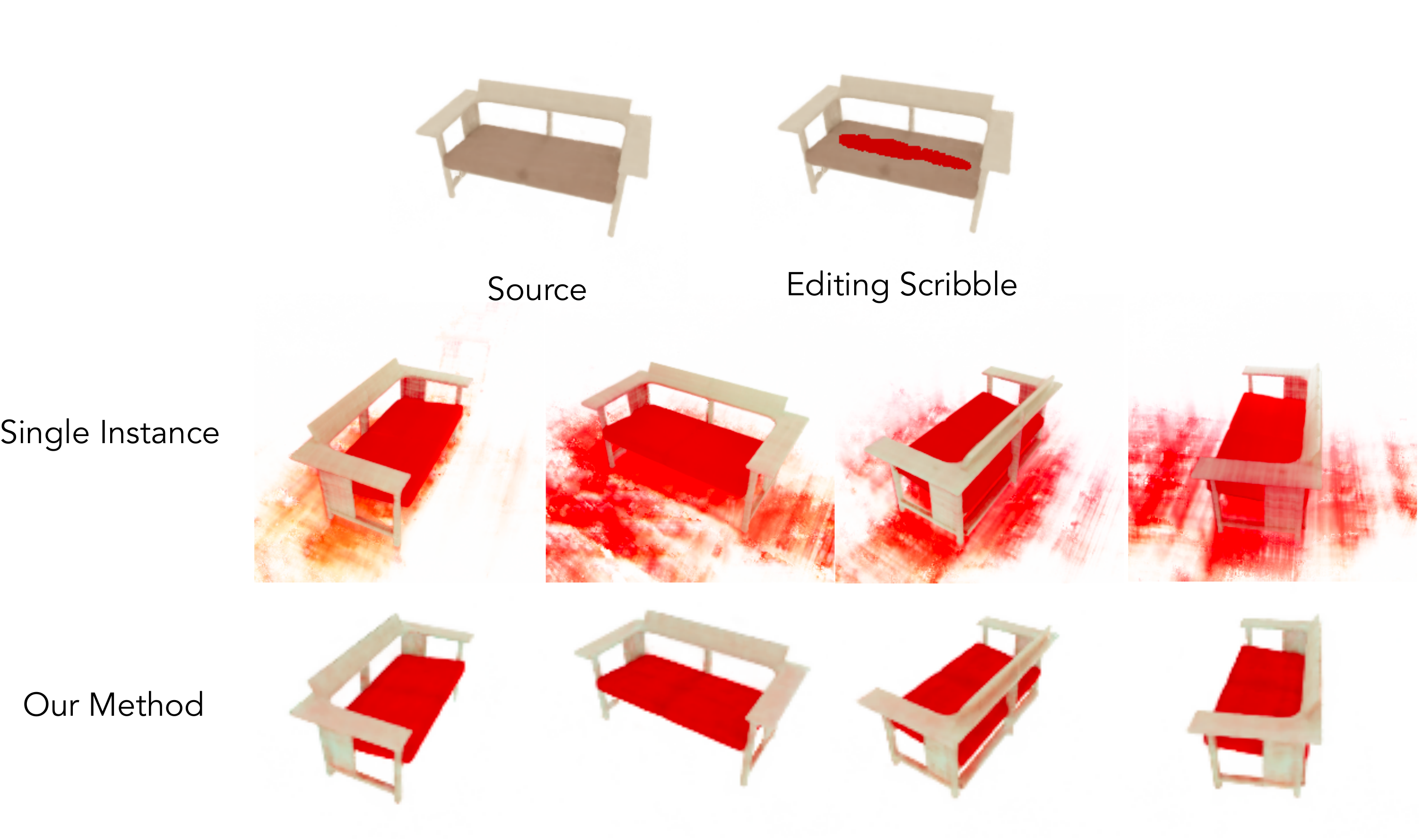}
  \caption{
  {\bf Single-instance vs.\ conditional radiance field editing.} 
  We visualize the edit of changing the color of a seat from brown to red. We find that edits on single-instance NeRFs propagate to outside the chair and cause artifacts in the background, whereas our model successfully propagates the edit across only the seat of the chair.
  }
  \lblfig{supp-singleinstance}
\end{figure*}

\myparagraph{Single-instance NeRF editing.}
We also provide an additional comparison of our method against editing a single-instance NeRF. Here, we change the color of a seat from brown to bright red. Again, we observe that the single-instance NeRF does not learn an edit that generalizes; the model frequently creates red artifacts in chair's background. In contrast, our model can still learn an edit that successfully propagates to the seat but not to other regions of the scene. We visualize these results in \reffig{supp-singleinstance}.

\begin{figure*}
  \centering
   \includegraphics[width=\linewidth]{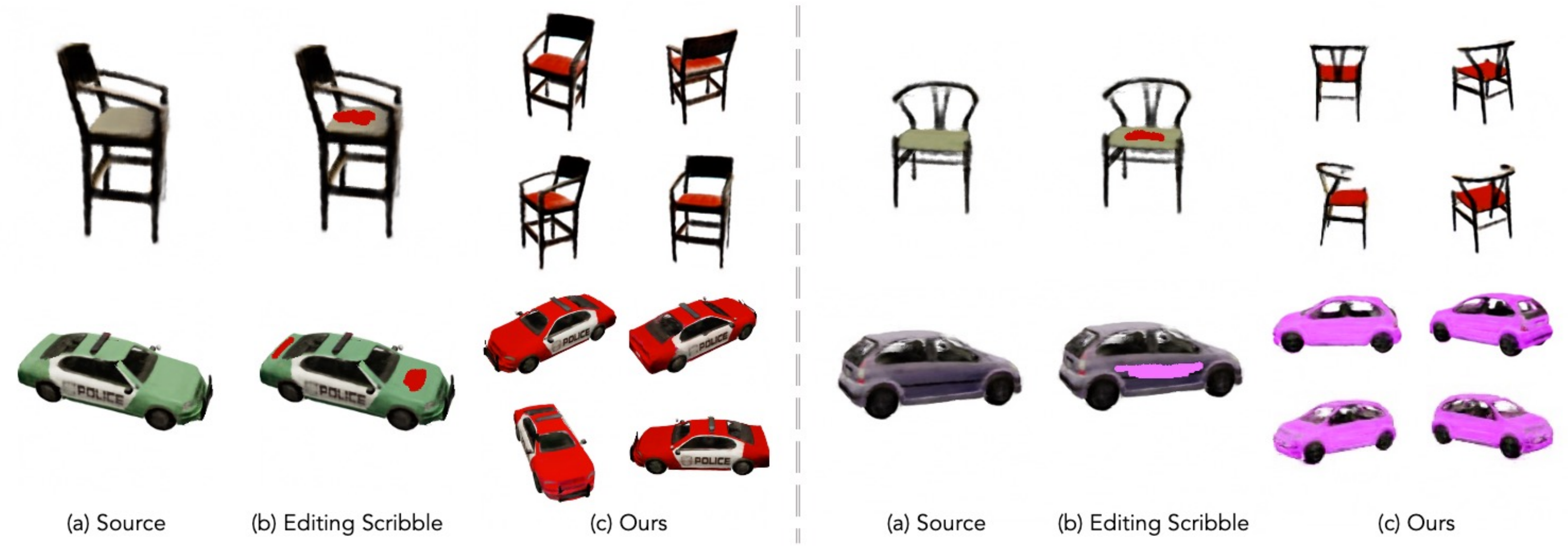}
  \caption{
  {\bf Additional color editing qualitative results.} 
  Our method successfully colors the seats of two Aubry \etal chairs~\cite{aubry2014seeing} to red, and changes the car body colors to red and pink. 
  }
  \lblfig{supp-figcoloreditsmore}
\end{figure*}

\myparagraph{Additional color editing results.}
We visualize additional color edits on the Aubry chairs~\cite{aubry2014seeing} and the CARLA cars~\cite{dosovitskiy2017carla, schwarz2020graf} datasets in~\reffig{supp-figcoloreditsmore}. 

\section{Additional Shape Edits}
\lblsec{additionalshape}
\begin{table*}
    \centering 
    \resizebox{\textwidth}{!}
    {
    \begin{tabular}{lrrrrrrrrr}
    \toprule
    \multicolumn{1}{c}{} &
    \multicolumn{3}{c}{Edit 1} & 
    \multicolumn{3}{c}{Edit 2} & 
    \multicolumn{3}{c}{Edit 3} \\ %
    \cmidrule(r){2-4}
    \cmidrule(r){5-7}
    \cmidrule(r){8-10}
            & \multicolumn{1}{c}{PSNR}  &  \multicolumn{1}{c}{SSIM}  & \multicolumn{1}{c}{LPIPS} 
            & \multicolumn{1}{c}{PSNR}  &  \multicolumn{1}{c}{SSIM}  & \multicolumn{1}{c}{LPIPS} 
            & \multicolumn{1}{c}{PSNR}  &  \multicolumn{1}{c}{SSIM}  & \multicolumn{1}{c}{LPIPS} \\
    \midrule
    Only Finetune Shape Code & $23.52$ & $0.947$ & $0.100$ & $20.18$ & $0.919$ & $0.138$ & $22.52$ & $0.927$ & $0.118$\\ %
    Only Shape Branch & $24.59$ & $0.947$ & $0.090$ & $17.96$ & $0.887$ & $0.160$ & $22.94$ & $0.929$ & $0.104$\\ %
    Finetuning All Weights & $21.31$ & $0.923$ & $0.100$ & $19.77$ & $0.903$ & $0.128$ & $19.84$ & $0.903$ & $0.123$ \\
    Our Method  & $\textbf{25.68}$ & $\textbf{0.958}$ & $\textbf{0.069}$ & $\textbf{22.97}$ & $\textbf{0.933}$ & $\textbf{0.091}$ & $\textbf{25.04}$ & $\textbf{0.943}$ & $\textbf{0.083}$  \\ %
    \bottomrule
    \end{tabular}
    }
    \caption{
    {\bf Shape editing quantitative results.} 
    We evaluate shape editing of a source object instance to match a target instance. Please refer to \reffig{supp-figshapeedits}  for a visualization of each of the edits. Notice that our method outperforms the baselines for all color edits on all criteria. }
    \lbltbl{supp-tblshapeedits}
\end{table*}
\begin{figure*}
  \centering
   \includegraphics[width=\linewidth]{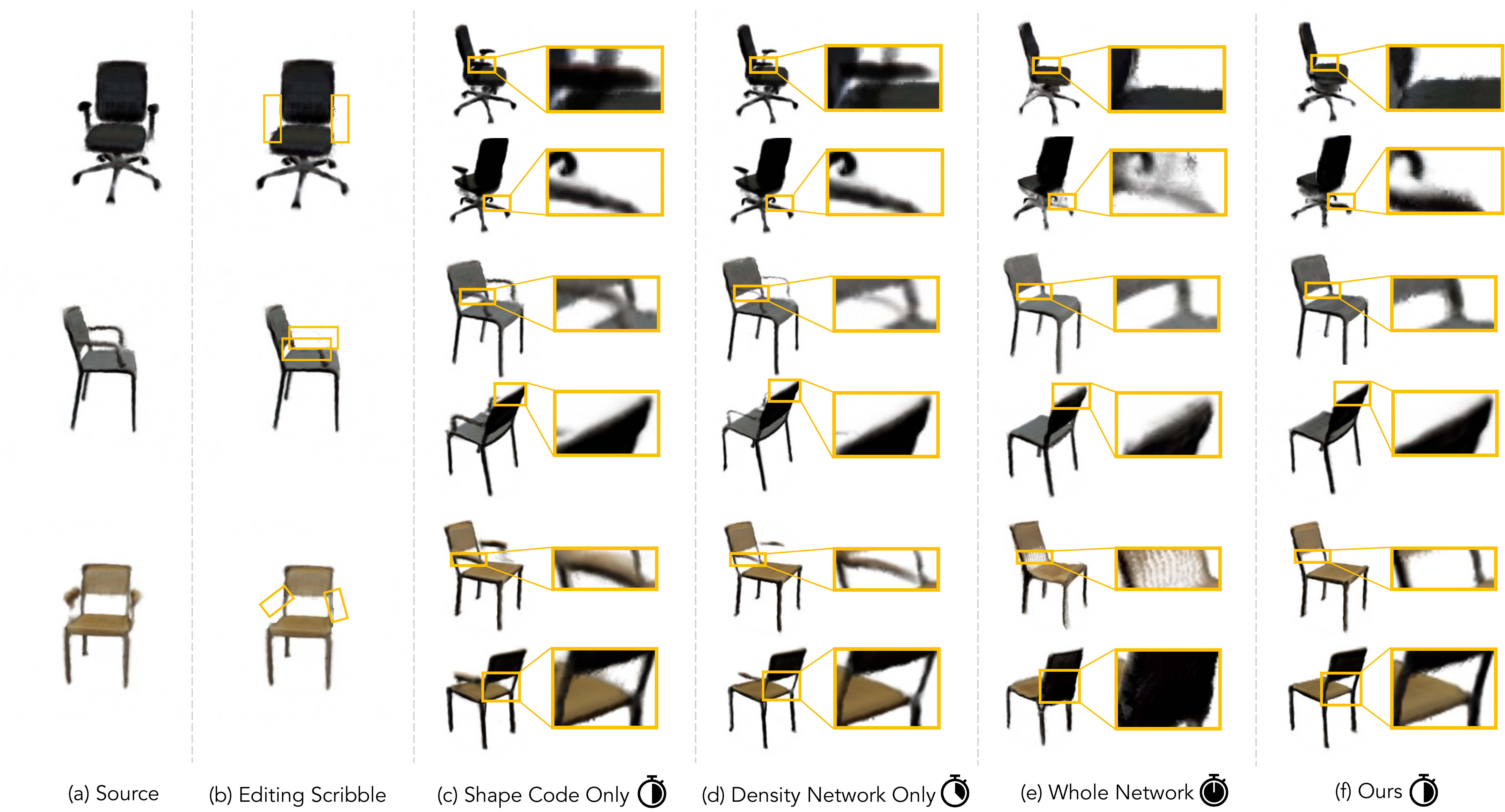}
  \caption{{\bf Shape editing quantitative results.} 
  Notice how only optimizing the shape code or branch are unable to fit both edits. Optimizing the whole network is slow and causes unwanted changes in the instance.}
  \lblfig{supp-figshapeedits}
\end{figure*}

\myparagraph{Quantitative color editing evaluation.}
\newtext{
In this section, we visualize all three shape edits used for evaluation in the main paper. We visualize the results of the three edits in \reffig{supp-figshapeedits} and quantify them in \reftbl{supp-tblshapeedits}. Visually, we see that consistent with the main paper, both finetuning the shape code and the shape branch are not enough to change the instance, but finetuning the whole network causes unwanted changes in the instance.}

\myparagraph{Single-instance NeRF editing.}
\begin{figure*}
  \centering
  \includegraphics[width=\linewidth]{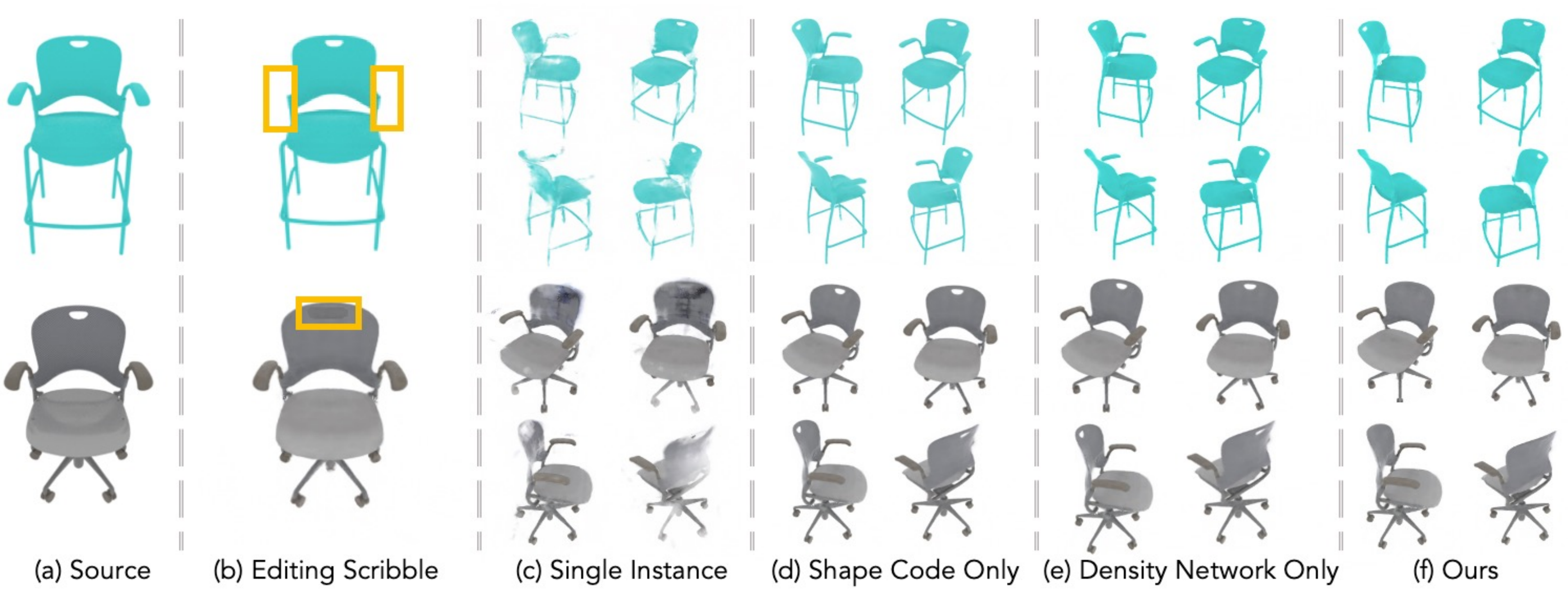}
  \caption{{\bf Shape editing qualitative results.} 
  Our method successfully removes the arms and fills in the hole of a chair. Notice how only optimizing the shape code or branch are unable to fit both edits. Furthermore, editing a single instance NeRF~\cite{mildenhall2020nerf} causes unwanted artifacts.}
  \lblfig{supp-shapeedits-photoshapes}
\end{figure*}
We  compare our method against editing a single-instance NeRF~\cite{mildenhall2020nerf}. We find that similar to the case with color edits, single-instance NeRFs are unable to learn an edit that generalizes to unseen views, likely due to a lack of a category-level prior. We visualize these results on the PhotoShapes dataset~\cite{park2018photoshape} in~\reffig{supp-shapeedits-photoshapes}. 

\begin{figure*}
  \centering
   \includegraphics[width=\linewidth]{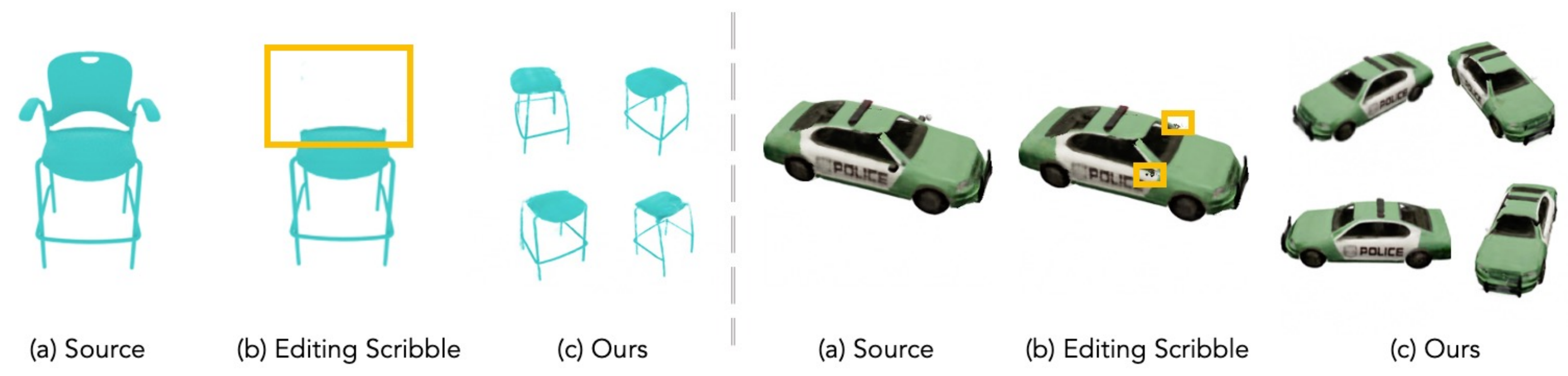}
  \caption{{\bf Shape editing qualitative results.} 
  Our method successfully removes the back and arms of a chair and removes the car mirrors.}
  \lblfig{supp-shapeeditsmore}
\end{figure*}

\myparagraph{Additional shape editing results.}
We visualize additional shape edits on the PhotoShapes~\cite{park2018photoshape} and the CARLA cars datasets~\cite{dosovitskiy2017carla, schwarz2020graf} in~\reffig{supp-shapeeditsmore}. 

\section{Additional Color/Shape Swapping Edits}
\lblsec{additionalswap}
\begin{figure*}
  \centering
   \includegraphics[width=\linewidth]{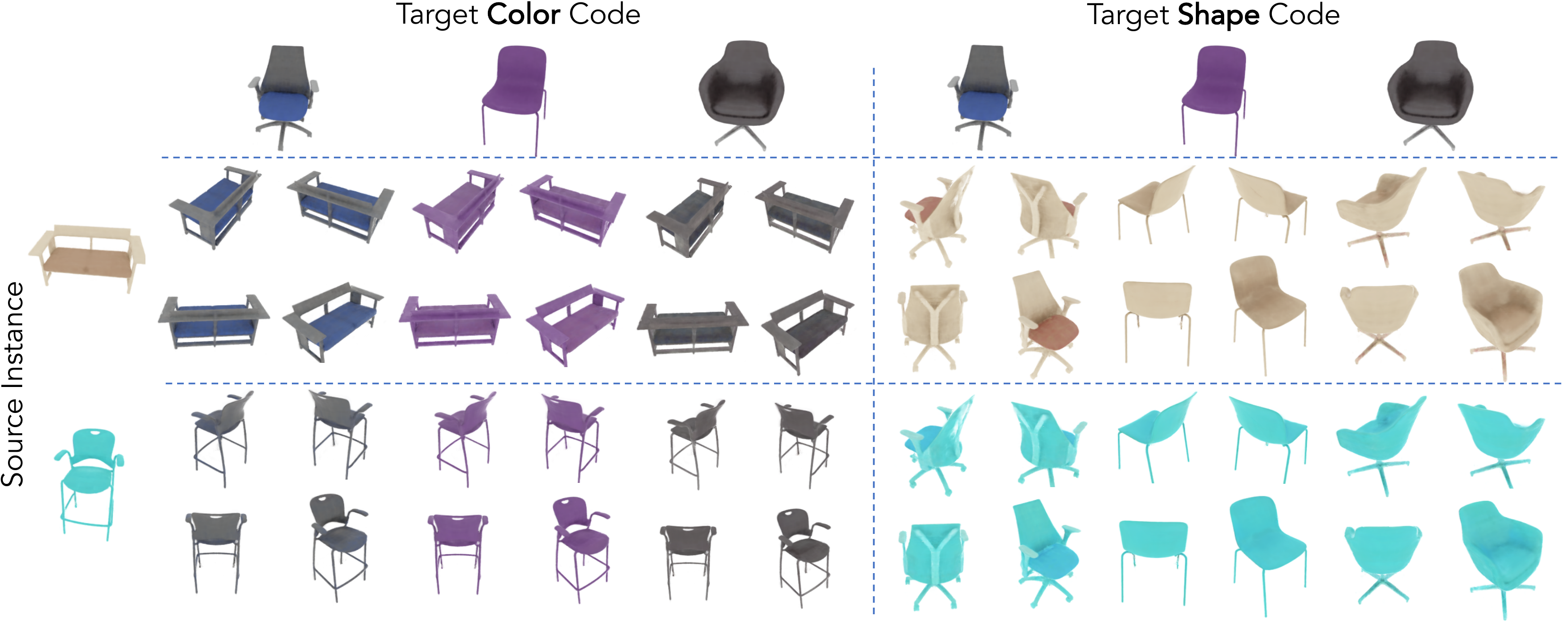}
  \caption{{\bf Shape and color transfer results.} Our model transfers the shape and color from target instances to a given source instance. Notice that when a source's color code is swapped with a target's, the shape remains unchanged, and vice versa.}
  \lblfig{supp-swapping}
\end{figure*}

We visualize additional shape and color swapping results on the PhotoShapes dataset~\cite{park2018photoshape} in~\reffig{supp-swapping} (this appendix). Notice again how changing the color code keeps the shape of the instance unchanged, and how changing the shape code keeps the color of the instance unchanged. 

\section{View Reconstruction}
\lblsec{reconstructions}
\begin{figure*}
  \centering
  \includegraphics[width=0.97\linewidth]{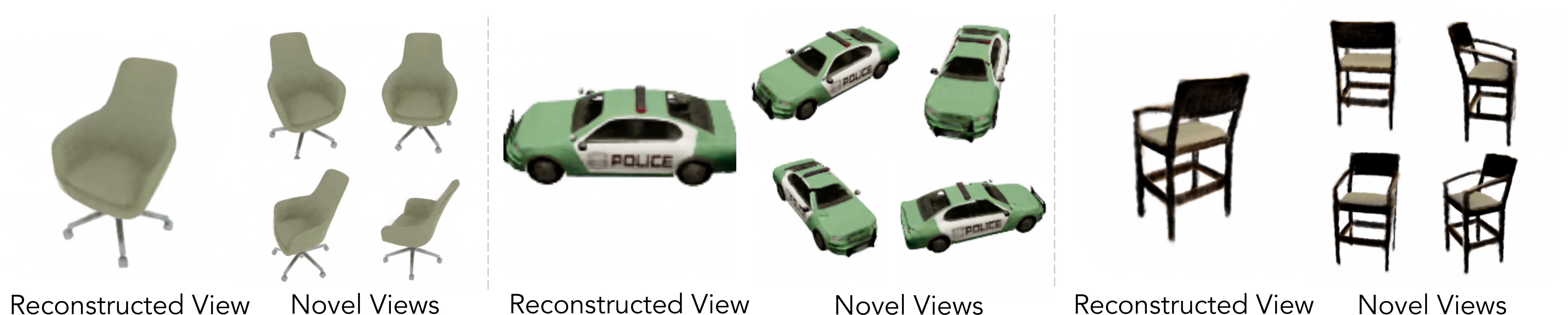}
\caption{ \newtext{
{\bf View reconstruction results.} Our method renders realistic and consistent views across several instances using a single model. }
}
  \lblfig{reconstructions}
\end{figure*}

\myparagraph{View consistency results.}
For each of our three datasets, we visualize synthesized views for a fixed instance and observe that the rendered views are all consistent in shape and color. We visualize these results in \reffig{reconstructions}. Notice how in the CARLA dataset~\cite{dosovitskiy2017carla, schwarz2020graf}, despite training on only one image per car instance, the model is able to infer the occluded regions of the car. 

\myparagraph{Additional reconstruction results.}
We visualize reconstructed views and depth maps across several instances of the PhotoShapes dataset~\cite{park2018photoshape} using our conditional radiance field. For each instance, we render four unseen viewpoints from our model and visually compare them against the ground truth views. We find that our method is able to almost perfectly reconstruct each instance, as well as learn convincing depth estimates of each instance.
We visualize reconstructions and depth maps for unseen views in Figures \ref{fig:supp-reconstructions12}-\ref{fig:supp-reconstructions910}.

\begin{figure*}
\centering
\includegraphics[width=0.97\textwidth, page=1]{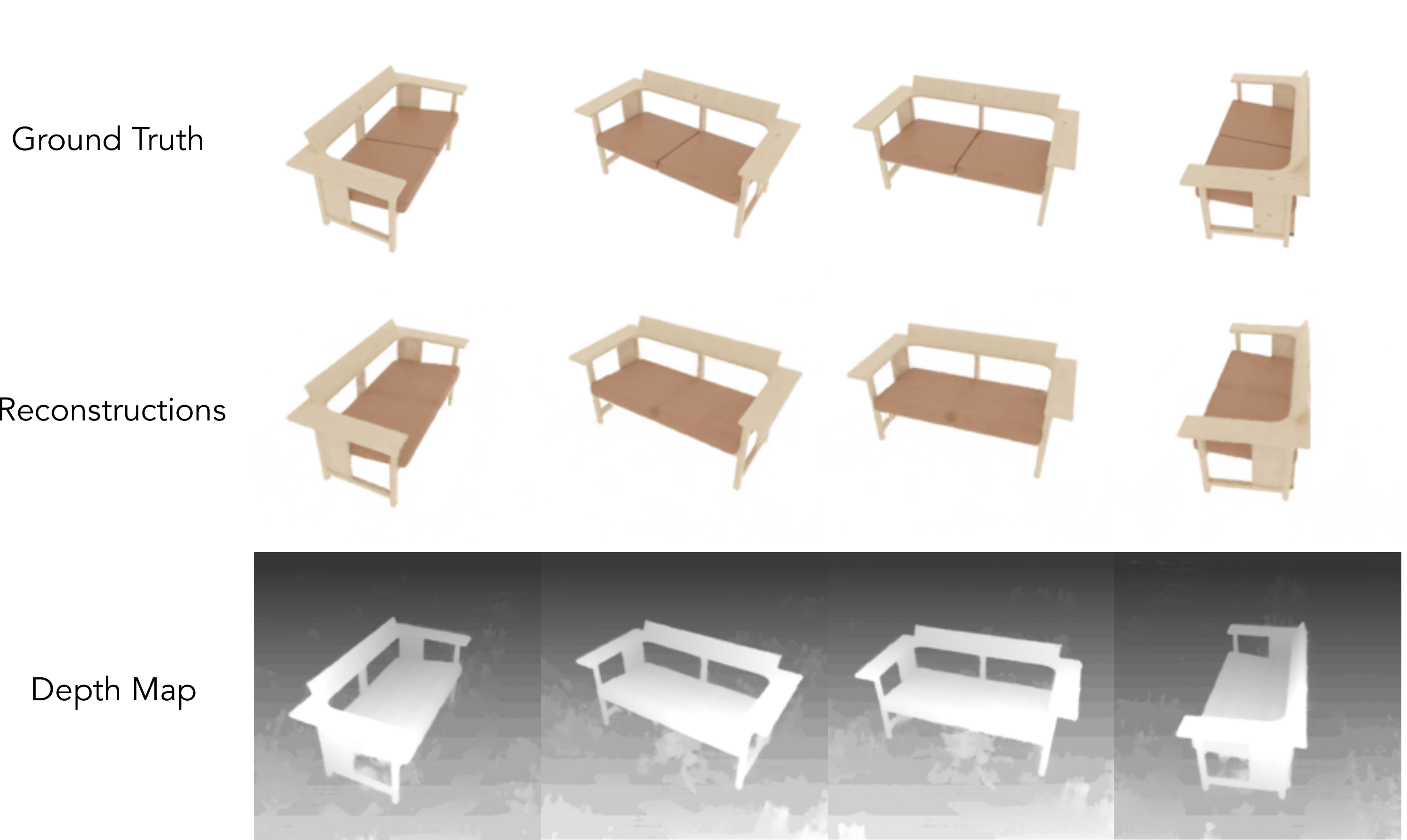}
\includegraphics[width=0.97\textwidth, page=2]{fig/reconstructions_all.pdf}
\caption{
  {\bf View reconstruction and depth prediction.} 
  We visualize the rendered views and predicted depth maps of our model on four unseen viewpoints. Notice our model is able to almost perfectly reconstruct the ground truth views.}
\lblfig{supp-reconstructions12}
\end{figure*}

\begin{figure*}
\centering
\includegraphics[width=0.97\textwidth, page=3]{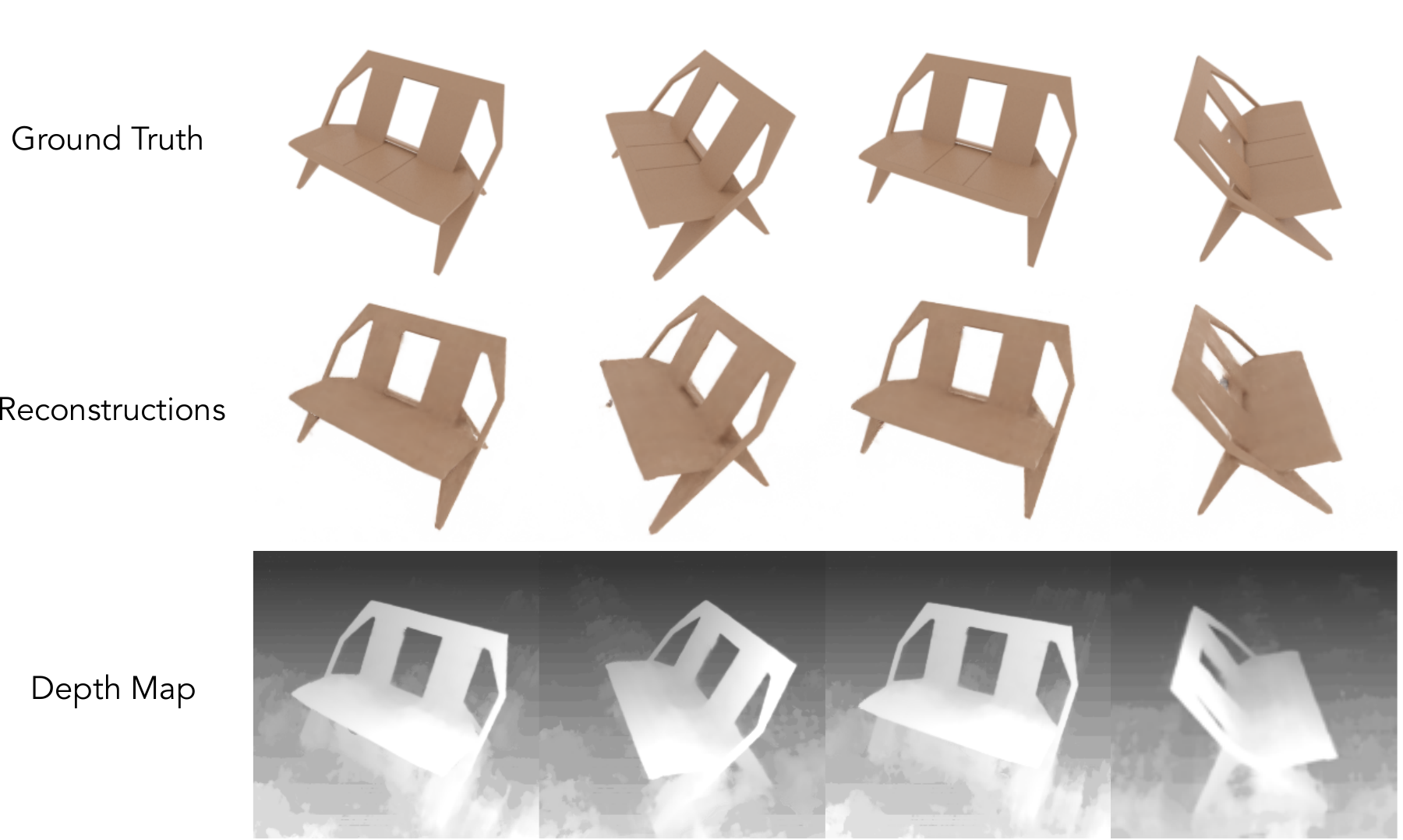}
\includegraphics[width=0.97\textwidth, page=4]{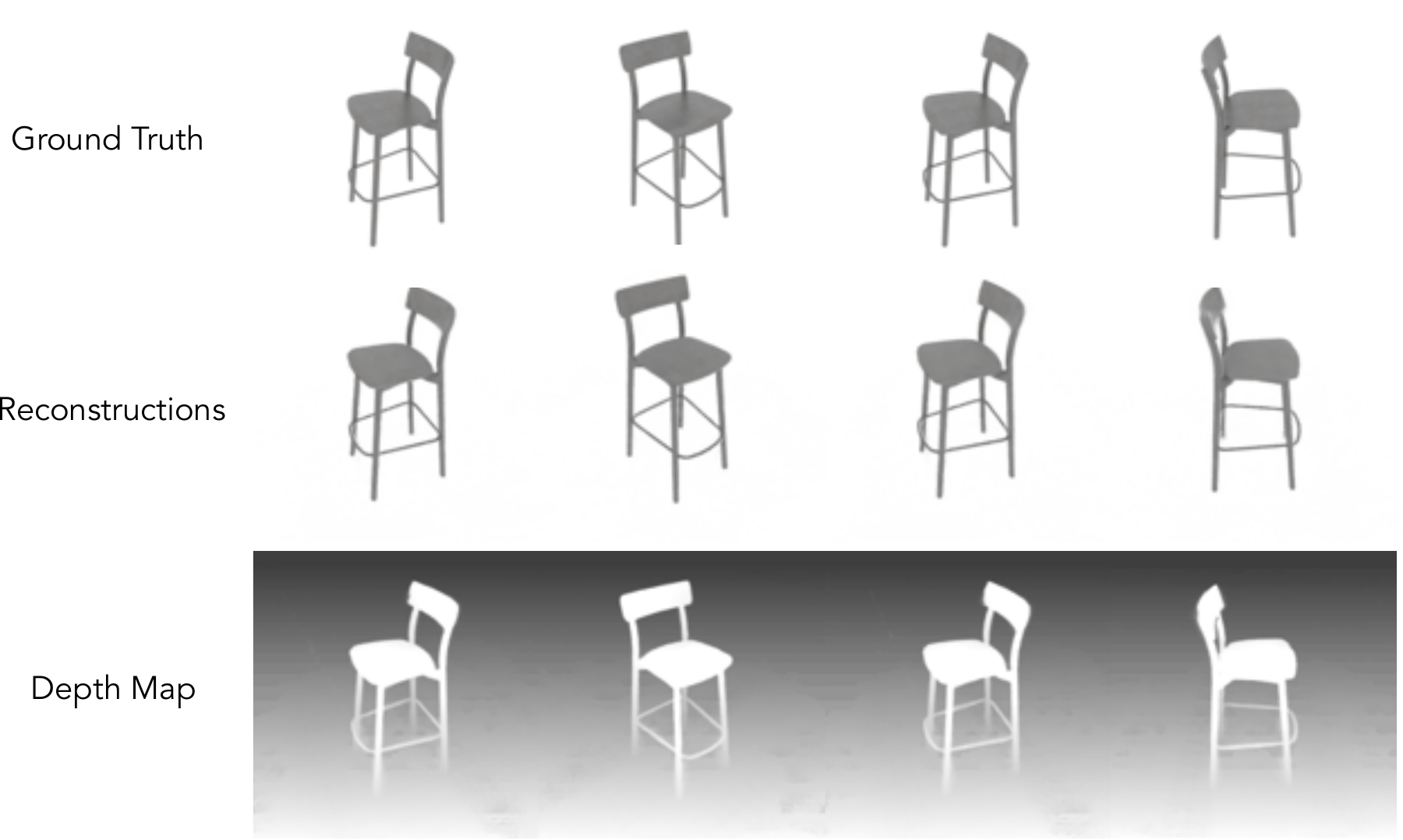}
\caption{
  {\bf View reconstruction and depth prediction.} 
  We visualize the rendered views and predicted depth maps of our model on four unseen viewpoints. Notice our model is able to almost perfectly reconstruct the ground truth views.}
\lblfig{supp-reconstructions34}
\end{figure*}

\begin{figure*}
\centering
\includegraphics[width=0.97\textwidth, page=5]{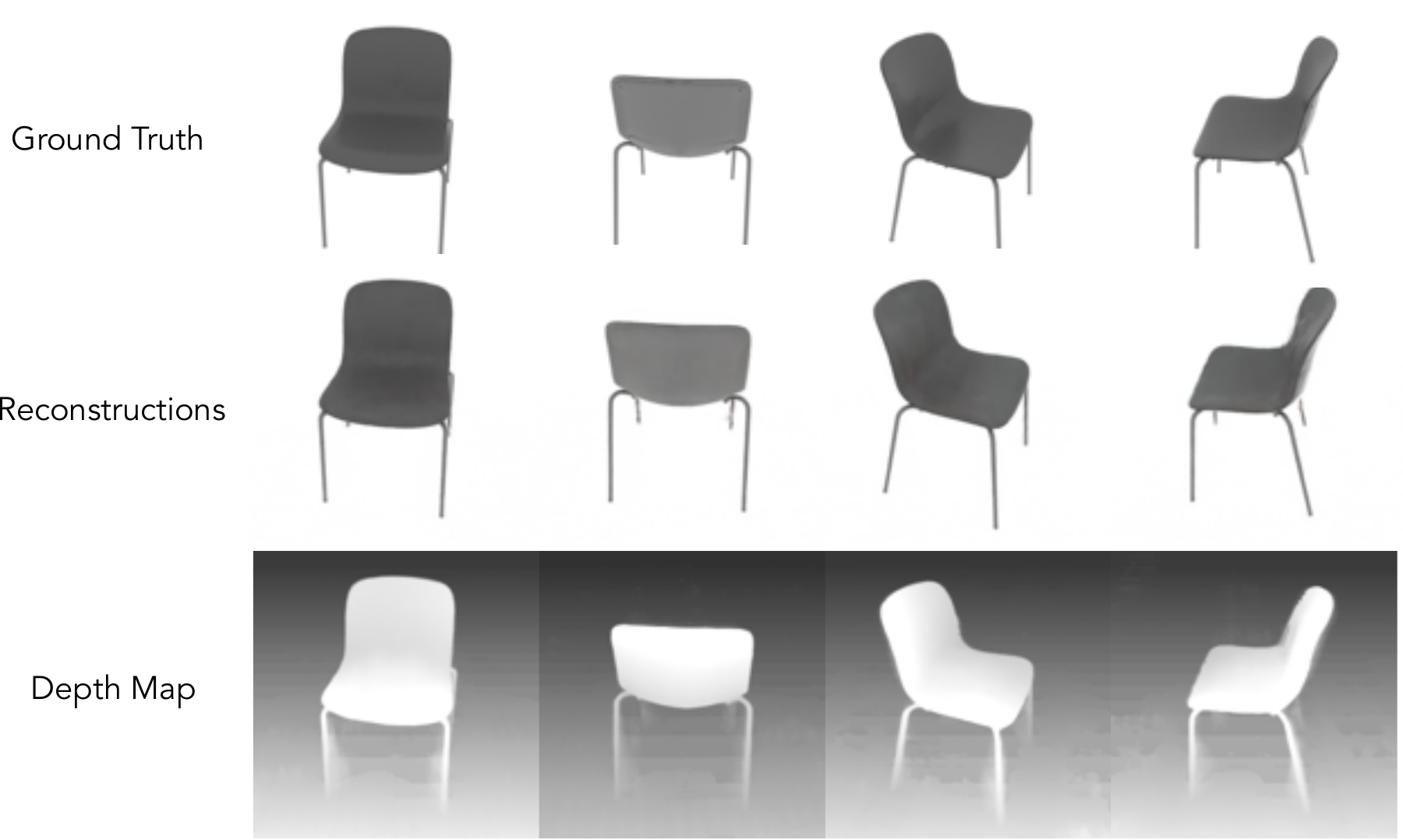}
\includegraphics[width=0.97\textwidth, page=6]{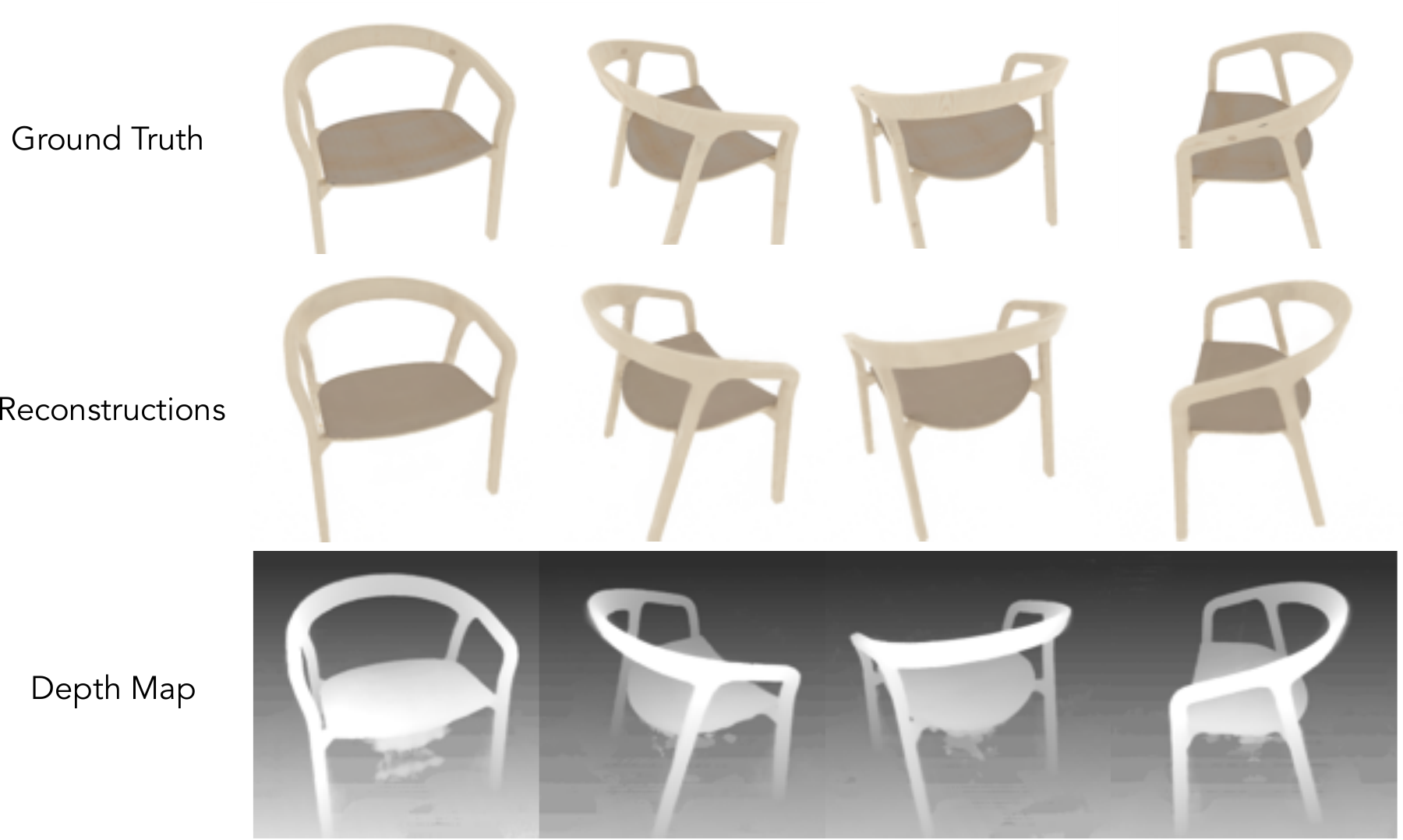}
\caption{
  {\bf View reconstruction and depth prediction.} 
  We visualize the rendered views and predicted depth maps of our model on four unseen viewpoints. Notice our model is able to almost perfectly reconstruct the ground truth views.}
\lblfig{supp-reconstructions56}
\end{figure*}

\begin{figure*}
\centering
\includegraphics[width=0.97\textwidth, page=7]{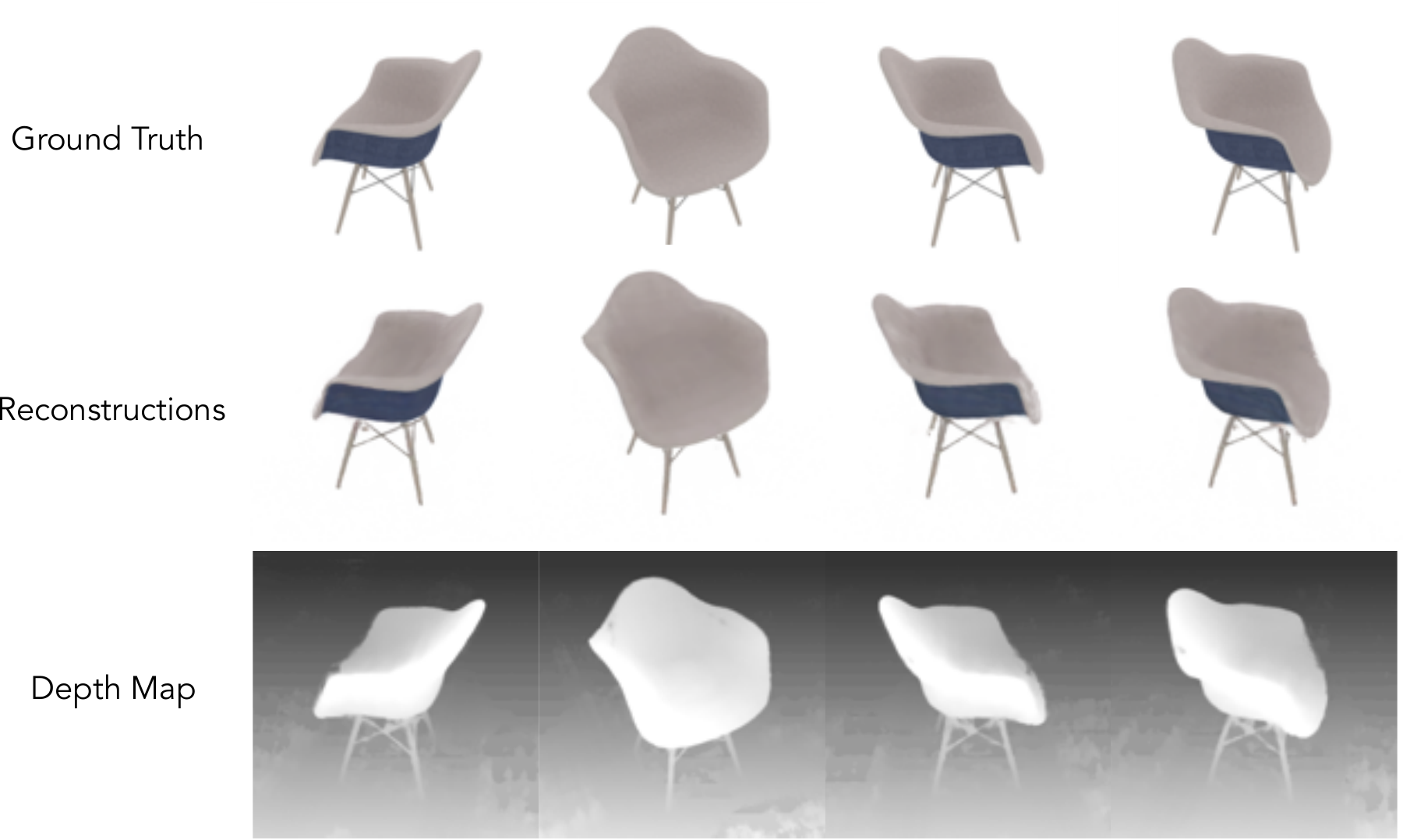}
\includegraphics[width=0.97\textwidth, page=8]{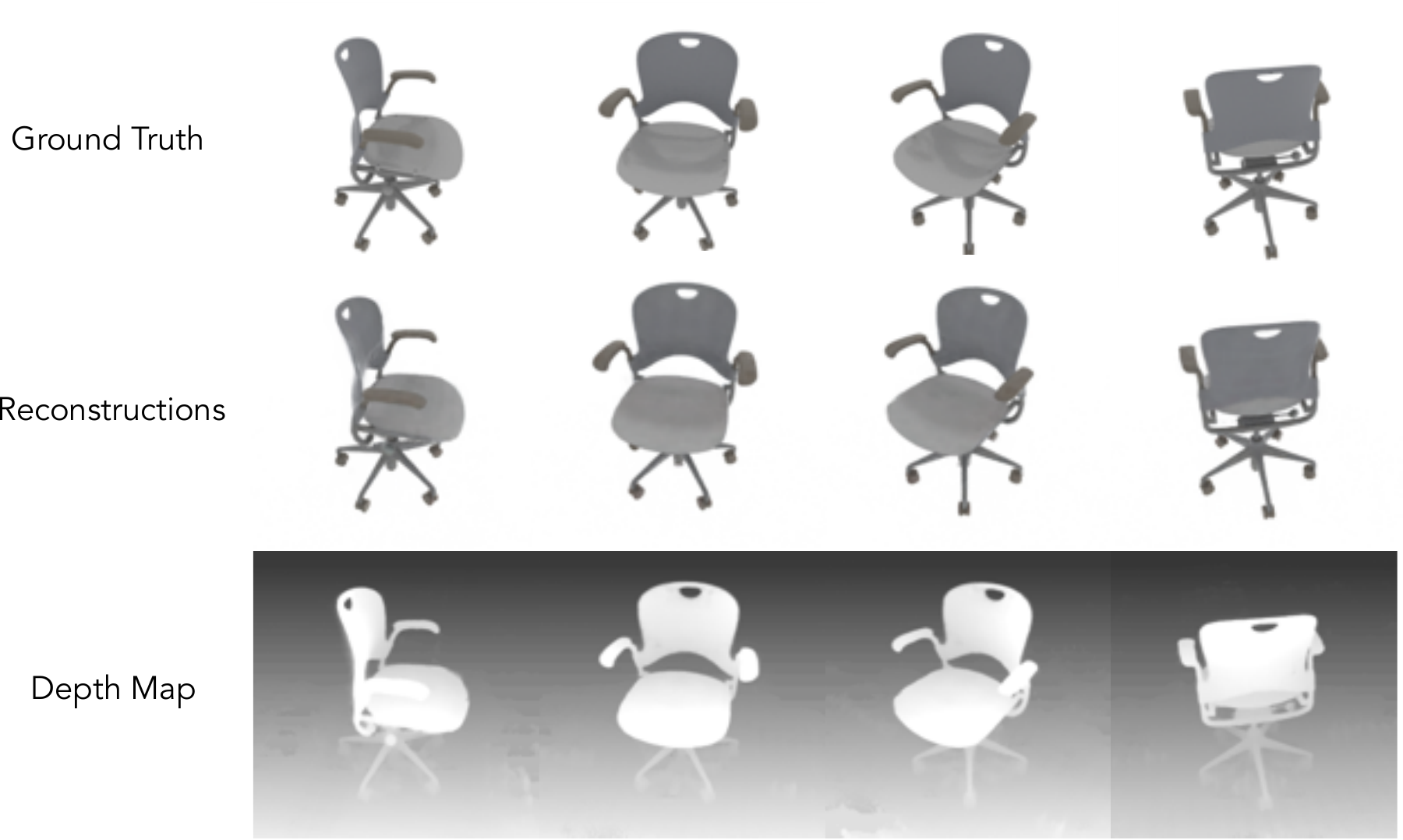}
\caption{
  {\bf View reconstruction and depth prediction.} 
  We visualize the rendered views and predicted depth maps of our model on four unseen viewpoints. Notice our model is able to almost perfectly reconstruct the ground truth views.}
\lblfig{supp-reconstructions78}
\end{figure*}

\begin{figure*}
\centering
\includegraphics[width=0.97\textwidth, page=9]{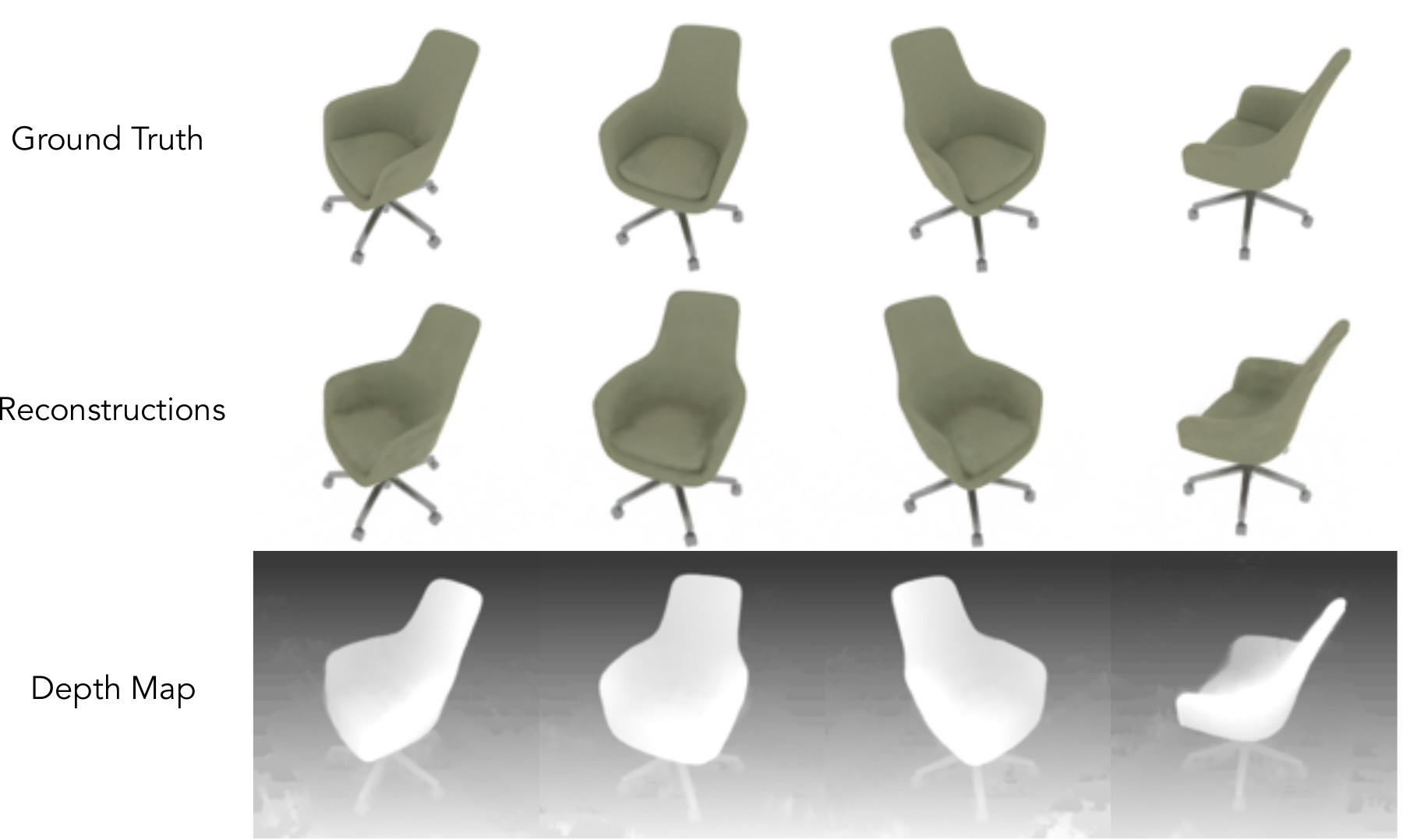}
\includegraphics[width=0.97\textwidth, page=10]{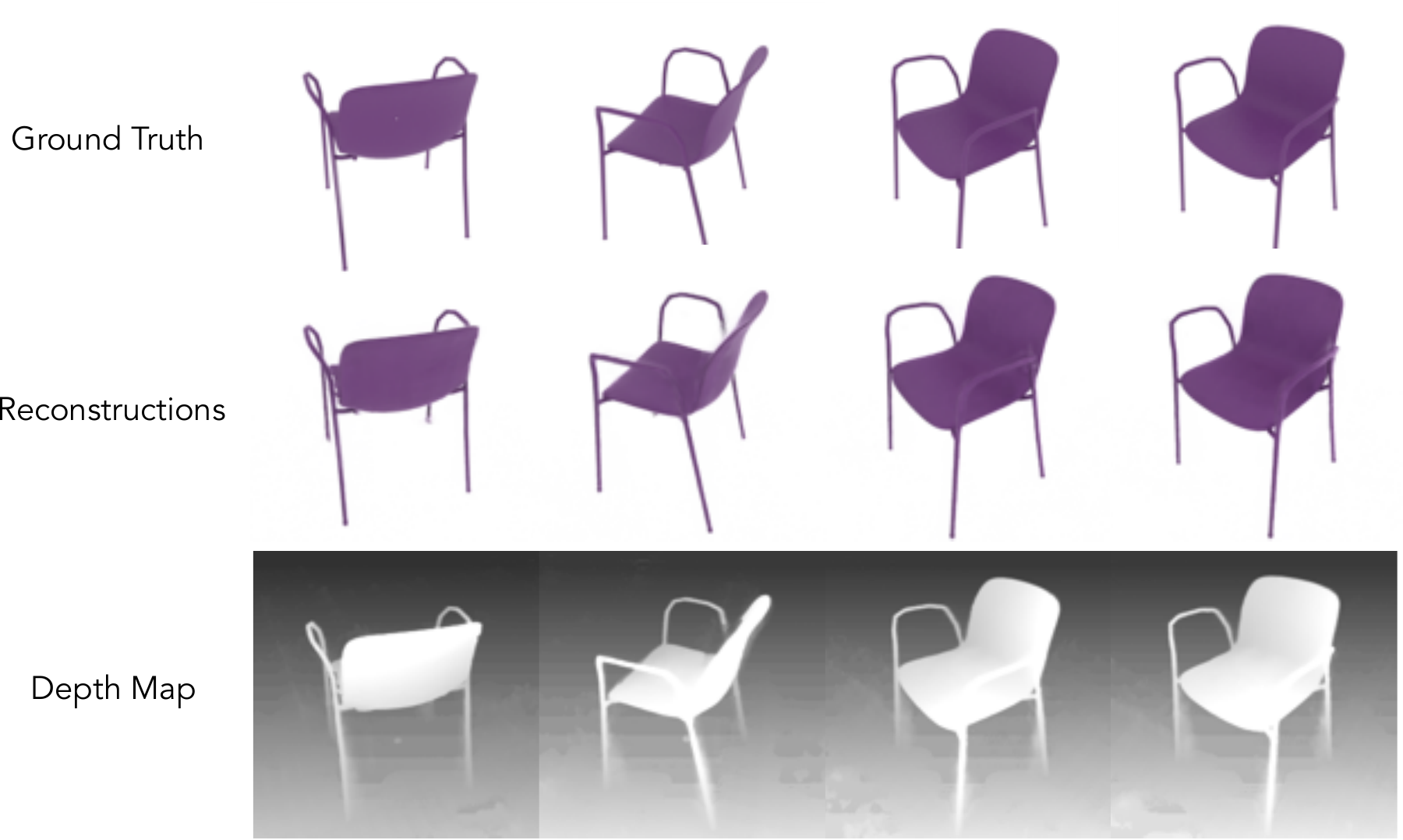}
\caption{
  {\bf View reconstruction and depth prediction.} 
  We visualize the rendered views and predicted depth maps of our model on four unseen viewpoints. Notice our model is able to almost perfectly reconstruct the ground truth views.}
\lblfig{supp-reconstructions910}
\end{figure*}

\section{Changelog}
\lblsec{changelog}
\myparagraph{v1} Initial preprint release.

\myparagraph{v2} Update \reffig{supp-figcoloredits} and include additional details in the appendix.

\end{document}